\documentclass[conference]{IEEEtran}

\let\labelindent\relax
\usepackage{enumitem}
\usepackage{times}
\usepackage[export]{adjustbox}
\usepackage{balance}

\usepackage[numbers]{natbib}
\usepackage{multicol}
\usepackage[bookmarks=true]{hyperref}
\usepackage{amsmath,amssymb}
\usepackage{xcolor}
\usepackage{graphicx}
\newtheorem{theorem}{Theorem}[section]
\usepackage{svg}
\usepackage{caption}
\usepackage{subcaption}
\usepackage{siunitx}
\usepackage{array,multirow}
\usepackage{booktabs}
\usepackage{graphicx}
\usepackage{makecell}
\usepackage{tabu}
\usepackage{soul}
\usepackage{kbordermatrix}
\newcommand{\todo}[1]{}
\usepackage[absolute,overlay]{textpos}

\newcommand{\methodname}{Causal MoMa}
\newcolumntype{P}[1]{>{\centering\arraybackslash}p{#1}}

\newtheorem{lemma}[theorem]{Lemma}
\newtheorem{prop}[theorem]{Proposition}

\begin{document}

% \begin{textblock*}{5in}(0.6in,0.32in) % {block width} (coords) 
% \textcolor{red}{\Large Draft compiled on \today}
% \end{textblock*}

\begin{textblock*}{7in}(0.7in,0.52in) % {block width} (coords) 
{\textbf{Robotics: Science and Systems 2023}}
\end{textblock*}

\title{Causal Policy Gradient for \\Whole-Body Mobile Manipulation}

% You will get a Paper-ID when submitting a pdf file to the conference system
% \author{Author Names Omitted for Anonymous Review. Paper-ID 260}

% \author{\authorblockN{Jiaheng Hu}
% \authorblockA{School of Electrical and\\Computer Engineering\\
% Georgia Institute of Technology\\
% Atlanta, Georgia 30332--0250\\
% Email: mshell@ece.gatech.edu}
% \and
% \authorblockN{Homer Simpson}
% \authorblockA{Twentieth Century Fox\\
% Email: homer@thesimpsons.com}
% \and
% \authorblockN{James Kirk\\ and Montgomery Scott}
% \authorblockA{Starfleet Academy\\
% San Francisco, California 96678-2391\\
% Telephone: (800) 555--1212\\

% avoiding spaces at the end of the author lines is not a problem with
% conference papers because we don't use \thanks or \IEEEmembership


\author{\authorblockN{Jiaheng Hu$^{1}$,
Peter Stone$^{1,2}$,
Roberto Martín-Martín$^{1}$
\authorblockA{$^{1}$The University of Texas at Austin $^{2}$SonyAI\\
\{jhu, pstone, robertomm\}@cs.utexas.edu}
}
}

\maketitle
% \thispagestyle{plain}
% \pagestyle{plain}

% \noindent\textcolor{red}{Note: Lots of polishing needed}\\
% \textcolor{red}{Deadline: February 3rd 23:59 AoE}\\

\begin{abstract}
Developing the next generation of household robot helpers requires combining locomotion and interaction capabilities, which is generally referred to as \emph{mobile manipulation} (MoMa). MoMa tasks are difficult due to the large action space of the robot and the common multi-objective nature of the task, e.g., efficiently reaching a goal while avoiding obstacles. Current approaches often segregate tasks into navigation without manipulation and stationary manipulation without locomotion by manually matching parts of the action space to MoMa sub-objectives (e.g. learning base actions for locomotion objectives and learning arm actions for manipulation). This solution prevents simultaneous combinations of locomotion and interaction degrees of freedom and requires human domain knowledge for both partitioning the action space and matching the action parts to the sub-objectives. 
In this paper, we introduce \methodname{}, a new reinforcement learning framework to train policies for typical MoMa tasks that makes use of the most favorable subspace of the robot's action space to address each sub-objective. \methodname{} automatically discovers the causal dependencies between actions and terms of the reward function and exploits these dependencies through causal policy gradient that reduces gradient variance compared to previous state-of-the-art reinforcement learning algorithms, improving convergence and results. We evaluate the performance of \methodname{} on three types of simulated robots across different MoMa tasks and demonstrate success in transferring the policies trained in simulation directly to a real robot, where our agent is able to follow moving goals and react to dynamic obstacles while simultaneously and synergistically controlling the whole-body: base, arm, and head. More information at \url{https://sites.google.com/view/causal-moma}
\end{abstract}

\IEEEpeerreviewmaketitle

\section{Introduction}
\label{s:intro}

\begin{figure}[t!]
\centering
\includegraphics[trim={0cm 5.5cm 18.2cm 0},clip,width=0.49\textwidth]{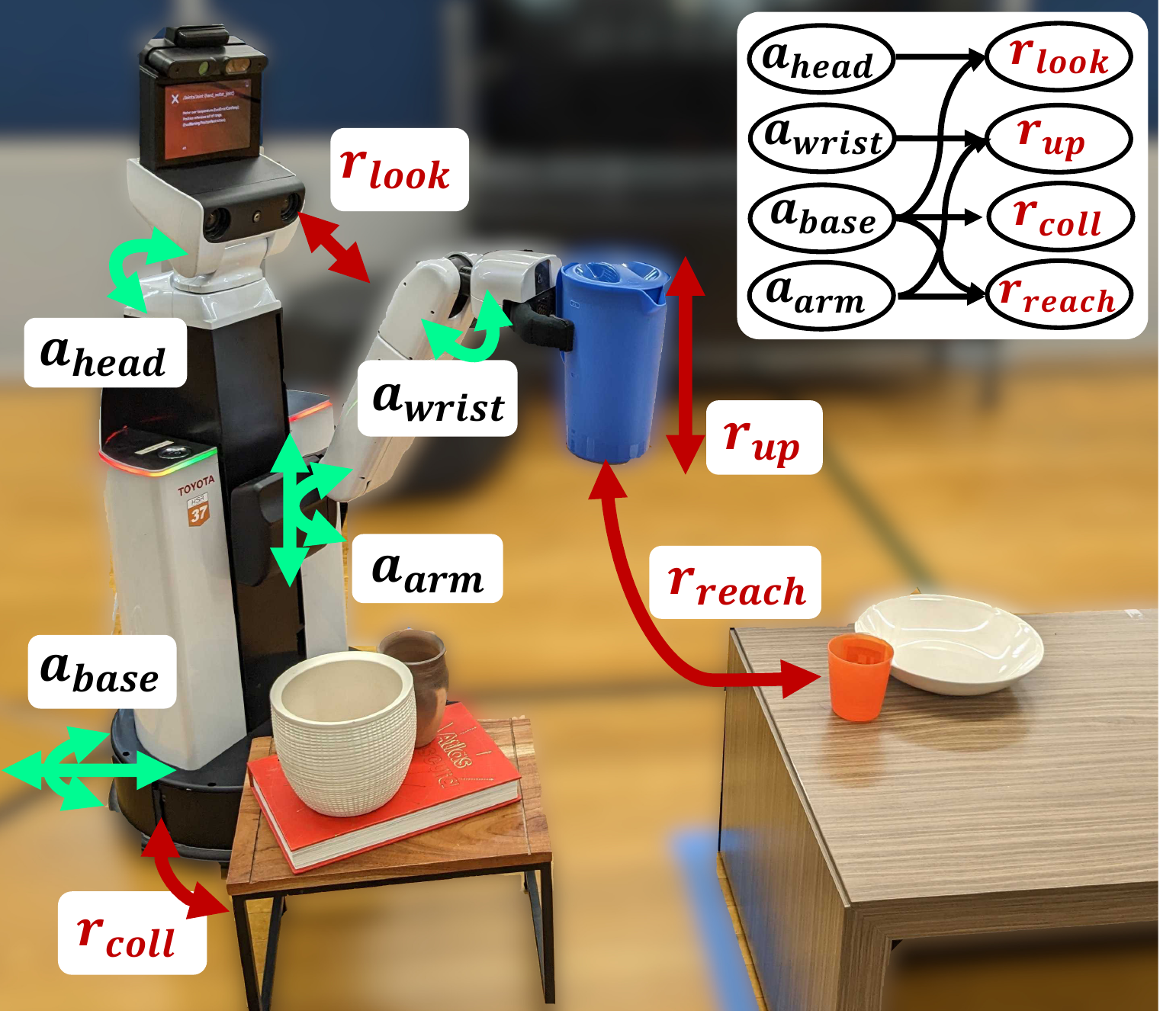}
\caption{Robot executing a mobile manipulation task: placing a jug on a table. The task is naturally defined by multiple objectives corresponding to a factored reward function with multiple components (red): reaching the placing location, keeping the orientation upright, looking at the goal, and avoiding collisions with the base. Only some subsets of the degrees of freedom of the robot (green) are necessary to fulfill each objective. This corresponds to causal dependencies between some action space dimensions and reward terms (top-right). \textit{\methodname{}} infers these underlying causal relationships and exploits them in a causal policy gradient approach that enables learning policies for complex mobile manipulation tasks.}
\label{fig:figure1}
\end{figure}

Mobile Manipulation (MoMa) requires combining locomotion and interaction capabilities for tasks that integrate elements from navigation and manipulation~\cite{siciliano2008springer,martin-martin_chalvatzaki_harada_2022}.
In sensorimotor control for navigation and for stationary manipulation, many of the most recent successes have come from posing the tasks as reinforcement learning (RL) problems~\cite{sutton2018reinforcement} and training a policy that maximizes the expected return defined by a reward function for the task, e.g., for stationary manipulation of rigid~\cite{levine2016end,popov2017data,rajeswaran2017learning,gu2017deep,nguyen2019review} and flexible objects~\cite{gupta2016learning,zhang2017toward,bhagat2019deep,matas2018sim}, and navigation based on visual data~\cite{zhu2017target,zhu2021deep,tai2017virtual,kulhanek2019vision,wijmans2019dd,perez2021robot}.
In robotic tasks, the reward function often takes on a composite form, where the eventual reward is a linear sum of a set of reward terms corresponding to a set of sub-objectives for the robot,
% The original problem becomes an optimization problem over the expected accumulated reward obtained from multiple reward functions defining multiple simultaneous objectives, 
e.g., navigating to a location (navigation sub-objective 1) without colliding with the environment (navigation sub-objective 2), or reaching a location with the end-effector (manipulation sub-objective 1) while maintaining it at a specific orientation (manipulation sub-objective 2).
While the composite reward can be a challenge for modern RL algorithms in stationary manipulation and navigation, it becomes insurmountable for RL in MoMa, where many of these sub-objectives are combined and must be optimized with a large action space resulting from the integration of locomotion and manipulation degrees of freedom (see Fig.~\ref{fig:figure1}). 

Our main insight is that the sensorimotor control learning in MoMa can be simplified and made tractable by finding and exploiting the existing strong correlation between parts of the controllable embodiment (i.e., dimensions of the action space) to each of the sub-objectives, i.e., elements of the reward signal. For example, collisions of the robot base with the environment are the result of wrong locomotion actions, independent of the arm movement, while the reason for a robot to collide with itself is usually the wrong use of arm commands, independent of the base actions. These strong causal dependencies need to be exploited to factorize and simplify MoMa reinforcement learning problems. 

In other domains, a priori known causal dependencies have been used to factorize and simplify RL problems via \textit{factored policy gradient}~\cite{spooner2021factored} or action-dependent factored baselines~\cite{wu2018variance}. 
The factorization reduces the variance on the common score-based gradient estimator used in RL that scales quadratically with the dimensionality of the action space.
But the factorization in prior work is the result of the researchers' manual hardcoding of task-domain knowledge, which limits the generalization and applicability of these methods.
In this work, we present \methodname{}, a two-step procedure to solve MoMa tasks with RL without a priori domain knowledge factorization: first, \methodname{} infers autonomously the causal dependencies existing between reward terms and action dimensions through a \textit{causal discovery}~\cite{tian2013causal} procedure. Then, \methodname{} leverages the discovered causal relationship within a policy learning solution based on \textit{causal policy gradients} that computes the advantage for each action based only on causally related reward terms. \methodname{}'s two-step procedure reduces the variance of policy gradient for MoMa tasks, achieving better performance compared to a set of baselines that include non-factored policy gradient and sampling-based motion planning. We evaluate the performance of our approach on two mobile manipulation domains with discrete and continuous action spaces, and with experiments on a real-world mobile manipulator, a Toyota HSR, and observe a significant improvement over the baselines.

In summary, in \methodname{} our contributions include:
\begin{itemize}[leftmargin=*]
    \item \methodname{}'s first step, a novel method to automatically discover the direct causal dependencies between controllable degrees of freedom of an agent (dimensions of the action space) and objectives (reward terms) in mobile manipulation tasks with composite reward,
    \item \methodname{}'s second step, the integration of the discovered causal dependencies into a policy learning implementation with causal policy gradients that reduces gradient variance and achieves superior performance compared to a state-of-the-art reinforcement learning algorithm (PPO) on mobile manipulation tasks,
    \item A demonstration of zero-shot transfer of the policies learned with \methodname{} from simulation onto a real-world mobile manipulator, and an empirical evaluation of the superior performance of our approach compared to a strong sampling-based planner (CBiRRT2) in the real world.
\end{itemize}

\begin{figure*}[t]
\centering
\includegraphics[trim={0cm 2cm 3cm 0},clip,width=0.99\textwidth]{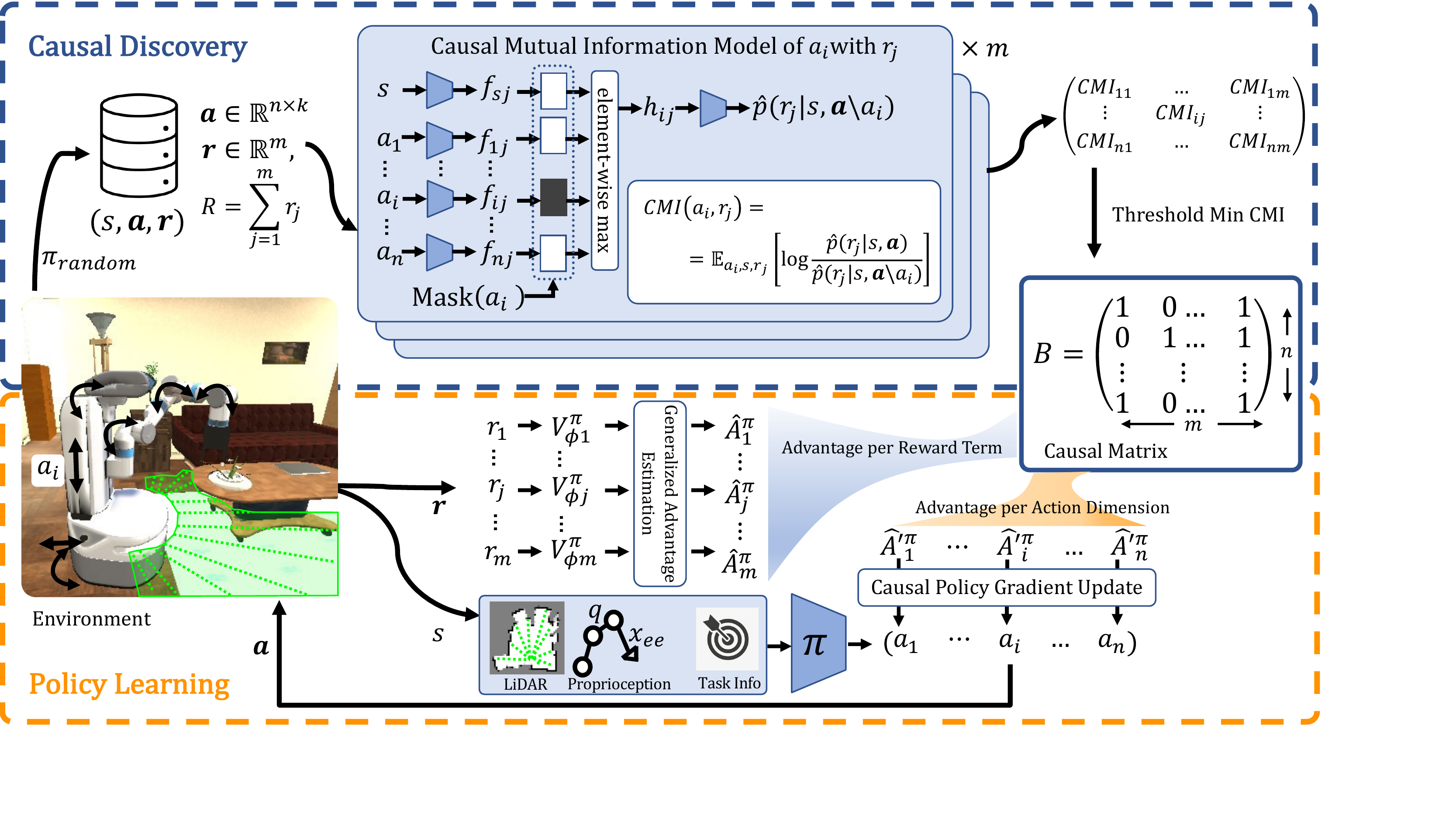}
% \includesvg[width=0.9\textwidth]{images/}
\caption{Two-step procedure in \textit{\methodname{}} for policy training in MoMa tasks with factored reward functions without a priori known action-space factorization. \textit{Top:} \methodname{} infers the causal dependencies existing between reward terms and action dimensions through a \textit{}{causal discovery procedure} on randomly collected data: estimating and thresholding the conditional-mutual information (CMI) between action dimensions and reward factors to infer the Causal Matrix, $B$. \textit{Bottom}: \methodname{} trains a policy that generates whole-body action commands based on onboard sensor signals and task information. For that, \methodname{} exploits the discovered Causal Matrix through causal policy gradient: advantages per reward term are aggregated into advantages for the causally related action dimension and used to update the policy, greatly reducing policy gradient variance.}
\label{fig:sysdiag}
\end{figure*}
% Removed the following
% in an unsupervised manner 

\section{Related Work}
\label{s:rw}

The prior research most relevant to ours comes from three general areas. First, since our target application is whole-body motion generation and control for MoMa, we summarize work in that area. Second, there is significant relevant past work on learning for mobile manipulators. Third, since the key motivation for factoring the action space is to reduce variance in the gradient estimation to improve RL policy learning, we review prior methods to reduce policy gradient variance.

\textbf{Generation of Whole-Body Motion for Mobile Manipulators:} Traditionally, the problem of coordinating locomotion and interaction in mobile manipulation has been explored through (whole-body) motion planning and control. When applying \textit{motion planning} to MoMa problems~\cite{stilman2010global,dai2014whole,burget2013whole, wolfe2010combined, kalakrishnan2011stomp}, uncertainty and inaccuracy in localization frequently impede the accurate execution of planned whole-body trajectories. In contrast, navigation and stationary manipulation are less sensitive to these inaccuracies due to the lower accuracy requirements of the former and the lack of base motion of the latter. As a result, researchers often factorize MoMa problems into sequences of navigation and stationary manipulation problems~\cite{stilman2007manipulation,xia2021relmogen,li2020hrl4in,kaelbling2013integrated}, losing the capabilities of synergistically combining all degrees of freedom.
Moreover, in order to use motion planning, the robot is typically assumed to have access to some form of geometric models for planning and localization during execution, and the environment is assumed static, a strong assumption in unstructured environments. 
When the robot task contains multiple sub-objectives, creating a whole-body motion planner is even harder, as it requires solving complex multi-objective optimization problems~\cite{huang2000coordinated,ratliff2009chomp,van2011lqg}. On the side of control, existing methods~\cite{seraji1998unified,yamamoto1992coordinating,siciliano2008springer,sentis2006whole,dietrich2012reactive,nori2015icub,papadopoulos2000planning, haviland2022holistic, pankert2020perceptive} resort to sophisticated prioritized solutions that require tuning to create controllers per objective and a prior decision by the developer on task priorities and necessary action dimensions for each task. These solutions cannot be guided by large dimensional sensor signals such as images or LiDAR scans. In comparison, \methodname{} learns a close-loop policy that simultaneously optimizes for multiple sub-objectives, is only based on onboard sensors, and does not require prior domain knowledge nor manual fine-tuning.

\textbf{Learning for Mobile Manipulation:}  Recently, reinforcement learning has shown to be a solution to overcome the aforementioned limitations of planning and control: it controls the robot based on onboard sensing, and the resulting controller policy is obtained autonomously from interactions with the environment instead of manually engineered. When applied to MoMa, previous works on reinforcement learning follow broadly two main strategies. The first group of methods uses a hierarchical controlling scheme with low-level policies that actuate predefined sections of the action space separately~\cite{li2020hrl4in, xia2021relmogen, ahn2022can, jauhri2022robot}. This approach allows the low-level policy to consider only a portion of the state and action space, enabling the system to train efficiently. However, by doing so, the robot forfeits the ability for whole-body motion. Our goal is to develop policies that can leverage the full potential of the whole body for MoMa tasks.

A second group of methods directly uses reinforcement learning over the entire action space to learn a policy. This enables the robot to utilize simultaneously all of its actuation capabilities but, when applied directly, is limited in the complexity of tasks and environments it can handle~\cite{wang2020learning}, with performance quickly deteriorating as tasks get more complex~\cite{kindle2020whole}. More complex MoMa tasks can be tackled by \citet{honerkamp2022n}, who learn base trajectories to adapt to given MoMa trajectories. However, they do not solve the original MoMa problem, only the base motion, and require a pre-computed obstacle map and a pre-defined trajectory cost function. \citet{fu2022deep} address full MoMa tasks by factorizing the action space into base and arm (as the hierarchical approaches) and assigning them manually to different sub-tasks. They compute separate RL advantages per body part and train a unified policy by mixing them. \methodname{} goes beyond theirs as it can mix advantages from an arbitrary number of action dimensions and reward terms, and does not need any manual assignation: \methodname{} discovers automatically the relation between action and reward. 
% \citet{honerkamp2022n} planned end-effector trajectories for MoMa using a sampling-based planner, then learned a base policy that adapts to the end-effector trajectory using reinforcement learning. Their method requires a pre-computed obstacle map and a pre-defined trajectory cost function. Our method, by contrast, directly learns a whole-body policy allowing the emergence of hard-to-engineer behaviors.
%\citet{gu2022multi} suggested the importance of having whole-body mobile manipulation skills for long-horizon task planning, and is orthogonal to our work, in that our method can improve the quality of the learned mobile manipulation skills. 
% We aim at combining the best of previous approaches in a general manner, enabling learning policies for MoMa through the discovery of causal relationship between action space dimensions and sub-objectives in the task without human intervention.

\textbf{Variance Reduction in Reinforcement Learning:} 
A fundamental problem in policy-based reinforcement learning is the large policy gradient variance, which leads to instabilities and failures in the training process~\cite{sutton2018reinforcement}. An effective method to reduce the variance without incurring bias is to use a baseline, typically in the form of a state-dependant value function~\cite{williams1992simple,schulman2015high}.
\citet{wu2018variance} studied using action-state-dependent baselines instead of state-dependent baselines to exploit the factorisability of the policy and further reduce variance. However, calculating an action-state-dependent baseline introduces additional computational overhead, e.g., additional neural network forward passes, which can quickly become a computational bottleneck. Furthermore, a closer examination by \citet{tucker2018mirage} showed empirically on multiple domains that learned state-action-dependent baselines do not reduce variance over state-dependent baselines.

For problems where there exist multiple objectives, \citet{spooner2021factored} showed that it is possible to derive a low variance state-dependent baseline by accounting for a known causal relationship between action dimensions and reward terms. Significantly, the factored policy gradient derived by \citet{spooner2021factored} can be computed efficiently with little computational overhead. \methodname{} adapts Factored Policy Gradient to the MoMa domain but does not require any given causal graph between action dimensions and objectives; it discovers it autonomously.

% \jeff{The following two paragraphs don't really fit into the section. Should we put them elsewhere or just delete them?}
% The idea of factorizing the advantage has also been explored in MARL literature. However the focus is on factorizing the value function, while the causal matrix is typically assumed to be an identity matrix (i.e. each agent is only responsible for its corresponding value)

% Most of the previous work in robot learning focuses on factorization of transition / dynamics. In our case, we are more interested in the influence of individual action dimensions on the components of a linear-additive reward function.

\section{\methodname}
\label{s:method}

We model the mobile manipulation task as a discrete-time Markov Decision Process represented by the tuple ($\mathcal{S}$, $\mathcal{A}$, $\mathcal{P}$, $R$, $\gamma$), where $\mathcal{S}$ is a state space, $\mathcal{A}$ is an action space, $\mathcal{P}$ is a Markovian transition model, and $R$ is a reward function. We study it as a reinforcement learning problem: the goal of our robot is to optimize the total expected return, characterized by a reward function $R(s, \mathbf{a})$ \footnote{For clarity of presentation, we indicate a full action vector with a bold $\mathbf{a}$ and a one-dimensional element with a non-bold $a$}. We assume that the robot's action is a $n$-dimensional vector $\mathbf{a}\in\mathcal{A}$ and that our reward function is factored, meaning that it is the linear sum of $m$ reward terms, i.e., $R(s, \mathbf{a}) = \sum_{j = 1}^{m} r_j(s, \mathbf{a})$.\footnote{One setting in which factored reward functions arise is when using reward shaping~\cite{ng1999policy} to densify an otherwise sparse-reward function; another common setting is when there are multiple objectives to accomplish.} Additionally, we define $\mathbf{r} $ as a vector with all the reward terms, i.e., $\mathbf{r} = (r_1,\ldots,r_{m})$. We note that the problem described above is a common setup for robotic tasks addressed with reinforcement learning, especially in mobile manipulation tasks where multiple objectives for navigation and manipulation represented by reward terms are combined.

Our main insight is to assume that, in MoMa tasks, only some dimensions of the action space are causally related to each term of the reward, i.e., for each $j\in\{1\dots m\}$, $a_i \rightarrow r_j$ only for some $i\in\{1\dots n\}$. By exploiting the intrinsic action-reward structure of MoMa problems, we can decompose the policy learning problem over the entire action space into a set of learning problems on subsets of the original action space, while still maintaining the ability to actuate all degrees of freedom simultaneously when necessary.
% In other words, for those pairs $p(r_j|a_i) $ \todo{finish this with the mathematical formalization of causal dependency}.

\methodname{} uses a two-step procedure to 1) discover this causal structure and 2) exploit it to train policies for MoMa tasks with factored reward, as depicted in Fig.~\ref{fig:sysdiag}. The steps include 1) an action-reward causal discovery procedure that infers autonomously the causal correlations between action dimensions and objectives (reward terms) from exploratory behavior, and 2) the integration of the learned causal structure into a causal policy gradient implementation obtained by modifying the proximal-policy optimization (PPO~\cite{schulman2017proximal}) algorithm. In the following paragraphs, we explain both steps in detail.

\subsection{Action-Reward Causal Discovery} 
\label{ss:arcd}

The first step of \methodname{} aims at inferring the causal relationship between action dimensions and reward terms, which can later be used to reduce policy gradient variance. 
We represent this relationship as a binary bi-adjacency $n\times m$ dimensional causal matrix $B$ that defines a bipartite causal graph (see Fig.~\ref{fig:sysdiag}, top). $B$ encodes the causal relation between action dimensions and reward terms, where $a_i \rightarrow r_j$ corresponds to $B_{ij} = 1$, and $a_i \not\to r_j$ to $B_{ij} = 0$.
During the causal discovery phase of our method, we infer the causal matrix ${B}$ from an exploratory dataset of robot actions collected via random interactions with the environment. 
Each data point in the exploratory dataset consists of a tuple $(s, \mathbf{a}, \mathbf{r})$, corresponding to the state, action, and vector of per-channel rewards at each timestep.
Our goal is to determine whether a causal edge $a_i \rightarrow r_j$ exists from each action dimension to each reward channel. We present a method for determining the existence of such causal relationships, based on the following \textbf{assumptions}:

\begin{itemize}[leftmargin=*]
  \item \textit{A1: Causal Markov Condition}~\cite{scheines1997introduction}. 
  Each variable in a causal graph, when conditioned on all its direct causes, is independent of all variables which are not effects or direct causes of it.
  \item \textit{A2: Faithfulness in action-reward correspondence}~\cite{scheines1997introduction}. There are no conditional independence relations other than the ones entailed by the Markov property. 
  \item \textit{A3: Uncorrelated action dimensions during exploration.} The exploratory data used for causal discovery has no correlation across action dimensions, i.e., for each pair of action dimensions, $i$ and $j$, at each timestep $t$, $p(a^t_i|a^t_j) = p(a^t_i)$. 
  \item \textit{A4: Action-reward causality inferable in short-horizon transitions.} The causal dependencies between action dimensions and reward terms can be observed in one or few timesteps, i.e., for every action-reward pair where a causal relation exists, the action will affect the reward within $k$ timesteps, where $k$ is a hyperparameter.
\end{itemize}
 
Both \textit{A1} and \textit{A2} are commonly made for causal inference.
We ensure \textit{A3} holds in our solution by collecting training data using a random policy. \textit{A4} typically holds for dense and semi-dense rewards (e.g. collision penalty) or reward terms that consist of potential / shaping components (e.g. goal position reaching), but may not hold in sparse reward settings. In this work, we use reward terms adapted from prior MoMa works~\cite{li2022igibson, xia2021relmogen, li2020hrl4in}, for which this assumption holds true.
We provide additional empirical analyses of \methodname{} in sparse reward settings in Appendix~\ref{app:sr}.
Notice that \textit{A4} does not impose that a longer time-horizon causal relation does not exist, as long as short time-horizon relation is also present. 

% \begin{theorem}
% \label{thm}
% Assuming \textit{A1}-\textit{A4}, and defining the conditioning set $\{s, \mathbf{a} \backslash a_i\}$ as $\{s, a_1, a_2, \ldots, a_{i-1}, a_{i+1}, \ldots, a_n \}$. We have $\forall i,j: $ $a_i \rightarrow r_j$ if and only if $a_i \not\!\perp\!\!\!\perp  r_j | \{s, \mathbf{a} \backslash a_i\}$.
% \end{theorem}

\begin{theorem}
\label{thm}
Let $\{s, \mathbf{a} \backslash a_i\}$ denote the conditioning set $\{s, a_1, a_2, \ldots, a_{i-1}, a_{i+1}, \ldots, a_n \}$. Let $\mathbf{a}^{t:t+k}$ denote a $n \times k$ dimensional matrix representing k-step actions from timestep $t$ to timestep $t+k-1$. 
Let $\mathbf{r}^{\sum{t:t+k}}$ denote a $m$ dimensional vector which is the vector sum of k-step rewards from timestep $t$ to timestep $t+k-1$.
Assuming \textit{A1}-\textit{A4}, we have \textit{$\forall i,j: $ $a_i \rightarrow r_j$ if and only if $a_i^{t:t+k} {\not\!\perp\!\!\!\perp}  r_j^{\sum{t:t+k}} | \{s^t, \mathbf{a}^{t:t+k} \backslash a_i^{t:t+k}\}$ for some timestep {t}. }
\end{theorem}

We provide the proof for this Theorem in Appendix~\ref{app:mc}. An intuitive way of understanding Theorem~\ref{thm} is that an action dimension $a_i$ is causally related to a reward term $r_j$ if and only if we find correlation between them within some k-step interval conditioning on all other action dimensions and the starting state of that interval.
Also notice that when k is set to 1, the correlation will be reduced to between two scalars, $a^t_i$ and $r_j^t$.
In the following paragraphs, we remove the timestep for simplicity and refer to $\mathbf{a}^{t:t+k}$ as $\mathbf{a}$, $\mathbf{r}^{\sum{t:t+k}}$ as $\mathbf{r}$, and $s^t$ as $s$.

Based on Theorem~\ref{thm}, the causal relationship between action dimensions and reward terms can be inferred through conditional independence tests, which can be made by measuring the Conditional Mutual Information (CMI) between pairs of action space dimensions and reward terms, $(i,j)$, as follows:
\begin{equation}
\begin{aligned} 
    \text{CMI}(a_i,r_j) &= \mathbb{E}_{a_i, s, r_j} \left[log\frac{p(a_i, r_j|\{s, \mathbf{a} \backslash a_i\})}{p(a_i|\{s, \mathbf{a} \backslash a_i\}) p(r_j|\{s, \mathbf{a} \backslash a_i\})}\right] \\&=\mathbb{E}_{a_i, s, r_j} \left[ log \frac{p(r_j | \{s, \mathbf{a} \})}{p(r_j | \{s, \mathbf{a} \backslash a_i\})}\right]
\end{aligned}
\end{equation}
where the expectation is taken over the joint distribution of $\{a_i, s, r_j\}$. We consider that a causal edge $a_i \rightarrow r_j$ exists if $\text{CMI}(a_i,r_j) > \epsilon$, where $\epsilon$ is a mutual-information threshold. 

We estimate the CMI between action dimensions and reward terms by training predictive models, $\hat{p}(r_j | \{s, \mathbf{a} \})$ and $\hat{p}(r_j | \{s, \mathbf{a} \backslash a_i\}$, for each $i, j$ over the exploratory dataset. However, training a separate model for estimating each probability would require a total of $n \times m$ models, which is computationally infeasible. To efficiently estimate the CMI between actions and rewards, we adopt the model architecture and training procedure proposed by \citet{wang2022causal} originally used for learning causal dynamic models, in the form explained below. This model architecture reduces the total number of models needed to $m$, the number of reward terms.

Specifically, for each reward channel $r_j$, we train a model (Fig.~\ref{fig:sysdiag}, top) that predicts the value of that reward channel from a full or partial action vector and the state, $\hat{p}(r_j | \{s, \mathbf{a} \})$ and $\hat{p}(r_j | \{s, \mathbf{a} \backslash a_i\}$. The model consists of three steps: first, each of the action dimensions, $a_1, ..., a_n$, and the state $s$ are individually mapped to feature vectors $f_{1,j}(a_1), ..., f_{n,j}(a_n), f_{s,j}(s)$ of equal length $l$ ($l=128$ in this work). Then, an overall feature $h_j$ is obtained by taking the element-wise max of all features. A prediction network $g_j()$ maps $h_j$ to the predicted reward channel. Using all values of the action vector as input, this procedure approximates the full conditional probability, $g_j(h_j)=\hat{p}(r_j | \{s, \mathbf{a} \})$. To estimate the conditional probability of the conditioning set for the action dimension $a_i$, we use a mask that sets the feature corresponding to $a_i$, $f_{i,j}(a_i)$, to $-\infty$ and repeat the reward inference obtaining $g_j(h_j)=\hat{p}(r_j | \{s, \mathbf{a} \backslash a_i\}$.
This model is trained to maximize the following log-likelihood:
\begin{equation}
\label{eqn:train_obj}
\begin{aligned} 
L = \sum_j [\log \hat{p}( r_j|s, \mathbf{a}) &+ \log \hat{p} (r_j | \{s, \mathbf{a} \backslash a_i\})]%\\ & + \log \hat{p}(r_j|P_{r_j} )]
\end{aligned}
\end{equation}
where $i$ is uniformly sampled from $\{1, \ldots, n \}$ for each $j$.
%, and $P_{r_j}$ are the parent nodes of the reward factor $r_j$ in the underlying causal graph, i.e., the states and actions that are causally related with $r_j$ inferred from the CMI based on the predicted rewards so far. We compute $P_{r_j}$ using the same reward prediction model from above, masking out all dimensions of the action vector and state that are considered independent. 
Maximizing equation~\ref{eqn:train_obj} corresponds to maximizing the accuracy of the two terms necessary for estimating $\text{CMI}(a_i,r_j)$, which promotes an accurate estimation of the causal graph. We split the exploration data into the training part for maximizing $L$ and the validation part for evaluating CMI.

After training, we obtain the bi-adjacency causal matrix ${B}$ by examining the CMI for each reward-action pair, $(i,j)$, based on the model's predicted conditional probability for each reward term in the validation dataset. Notice that this causal inference step incurs both data and computational overhead; we consider this impact in our evaluation in Sec.~\ref{s:exp}.

\subsection{Policy Learning}
Once \methodname{} has inferred the causal matrix $B$ through causal discovery, it uses it to reduce the policy gradient variance and learn a MoMa policy with a modified policy gradient procedure (see Fig.~\ref{fig:sysdiag}, bottom). For this purpose, we redefine the policy gradient to be
\begin{equation}
    \begin{aligned}
        \nabla_{\theta} J(\theta) = \nabla_{\theta} \log \pi_{\theta} (\mathbf{a} | s) \cdot {B} \cdot \hat{\mathbb{A}}^\pi(s, \mathbf{a}) 
    \end{aligned}
    \label{eqn:fpg}
\end{equation}
where $\theta$ is the policy parameter, $\hat{\mathbb{A}}^\pi(s, \mathbf{a})$ is a $m$-dimensional vector representing the advantage function factored across the reward terms, $\hat{\mathbb{A}}^\pi(s, \mathbf{a})=(\hat{\mathbb{A}}^\pi_1,\dots,\hat{\mathbb{A}}^\pi_m)$, and $\nabla_{\theta} \log \pi_{\theta} (\mathbf{a} | s)$ is a $|\theta| \times n$ matrix with each column corresponding to the log gradient of a particular action dimension's probability.\footnote{Notice that the $\nabla_{\theta} \log \pi_{\theta} (a | s)$ here is different from the canonical notation, which typically refers to a $|\theta|$-dimensional vector representing the log gradient of the entire action vector's probability}

\begin{theorem}
\label{thm_fpg}
\textit{If the causal matrix $B$ is correct, then Equation~\ref{eqn:fpg} is an unbiased estimator of the true policy gradient.}
\end{theorem}

We include proof of this Theorem in Appendix~\ref{ss:prof_cpg}.
Theorem~\ref{thm_fpg} entails that we can use the modified version of the policy gradient while keeping the original convergence guarantees. Moreover, \citet{spooner2021factored} demonstrated that the variance reduction of Equation~\ref{eqn:fpg} compared to the original policy gradient will be non-negative, and is often significant when the causal matrix $B$ is sparse.
An intuitive way of understanding the variance reduction here is to note that for each action dimension $a_i$, reward terms that $a_i$ cannot affect will only contribute noise to its update. By multiplying the per-reward-advantage $\hat{\mathbb{A}}^\pi(s, \mathbf{a})$ with the causal matrix $B$, we actively remove the irrelevant terms from the policy gradient estimator.
Through this approach, \methodname{} actively reduces gradient variance, stabilizing the training process while retaining the original theoretical guarantees, all without losing any of the capabilities of the agent to combine locomotion and arm(s) interactions when necessary. 

While our framework can be applied to any policy gradient and actor-critic algorithms, in this work we focus on using Proximal Policy Optimization (PPO)~\cite{schulman2017proximal} due to its simplicity and stability. Specifically, we modify the value network, $V_{\phi}$, to be multi-dimensional such that $V_{\phi}^{\pi}(s)\in\mathbb{R}^{m}$, one value for each of the $m$ reward terms. 
During \methodname{} policy training (Fig.~\ref{fig:sysdiag}, bottom), the agent takes actions $\mathbf{a} \sim \pi_\theta(\mathbf{a} | s)$ in the environment generating tuples $(s, \mathbf{a}, s', \mathbf{r})$, where $\mathbf{r}\in\mathbb{R}^{m}$ with $r_j$ being each of the reward terms. The value network is then updated using the target $\mathbf{r} + \gamma V_{\phi}^{\pi}(s')$. \methodname{} then calculates the per-reward-channel advantage $\hat{\mathbb{A}}^\pi(s, \mathbf{a})$ using Generalized Advantage Estimate~\cite{schulman2015high}, and obtain the per-action-dimension advantage with $\hat{\mathbb{A'}}^\pi(s, \mathbf{a}) = {B} \cdot \hat{\mathbb{A}}^\pi(s, \mathbf{a}) $. Lastly, \methodname{} updates its policy network $\pi_\theta(s)$ with causal policy gradient updates: updating each action dimension separately using the per-action-dimension advantage $\hat{\mathbb{A}'}(s, \mathbf{a})$ with the PPO policy objective.

% \begin{equation}
%     \begin{aligned}
%         \nabla_{\theta} J(\theta) = \hat{\mathbb{A}}(s, \mathbf{a}) \cdot {B} \cdot\nabla_{\theta} \log \pi_{\theta} (s | a)
%     \end{aligned}
%     \label{eqn:fpg2}
% \end{equation}

\begin{figure}[t]
\centering
\includegraphics[width=0.16\textwidth,height=0.23\textwidth,valign=t]{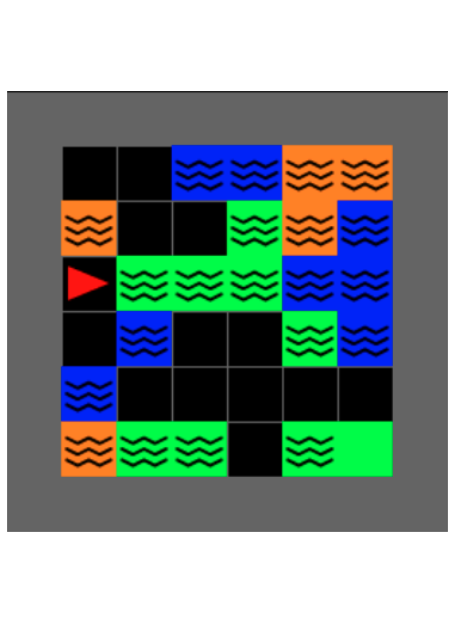}%{images/minipad.png}
\hfill
\includegraphics[width=0.62\columnwidth,valign=t]{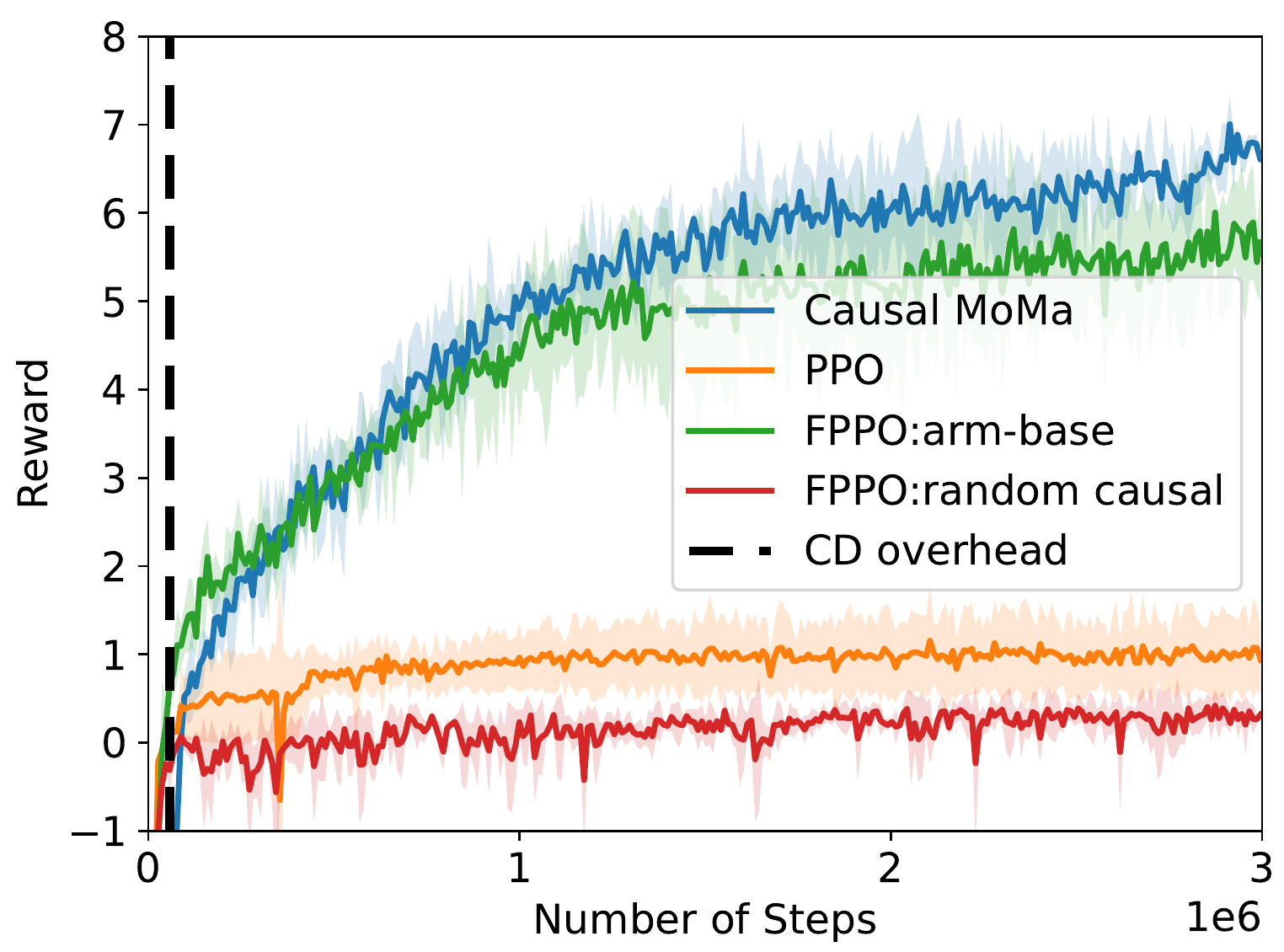}
% \begin{subfigure}[b]{1.05\columnwidth}
% \includegraphics[width=\linewidth]{images/minigrid_result_shifted.png}
% \caption{}
% \end{subfigure}
\caption{Experimental evaluation of \methodname{} on the Minigrid~\cite{minigrid} domain: (\textit{Left}) the agent controls an embodiment (red triangle) with discrete actions for navigation and manipulation with the goal of reaching a goal location (green tile). Blue tiles and green tiles require the agent to perform specific virtual manipulation actions. Orange tiles should be avoided by the agent. (\textit{Right}) training curves for \methodname{} and baselines, five seeds each, mean and std: \methodname{} converges to the highest reward thanks to the discovery and exploitation of the causal dependencies between actions and reward terms.}
\label{fig:minigridexp}
\end{figure}

\begin{figure*}[t]
\centering
\includegraphics[width=0.3\textwidth, height=0.22\textwidth]{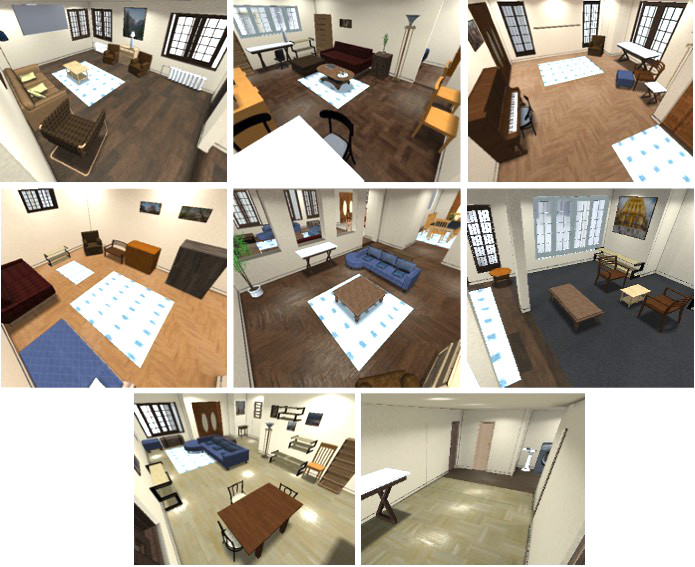}\hfill
\begin{subfigure}[b]{0.09\textwidth}
\centering
\includegraphics[width=0.5\textwidth]{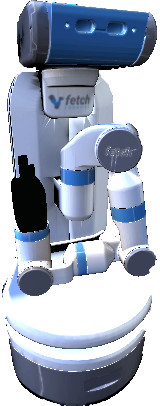}\\
\includegraphics[width=0.6\textwidth]{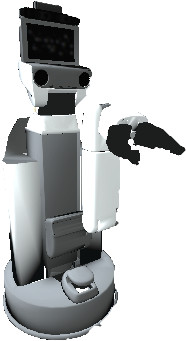}\hfill
\end{subfigure}
\includegraphics[width=0.3\textwidth]{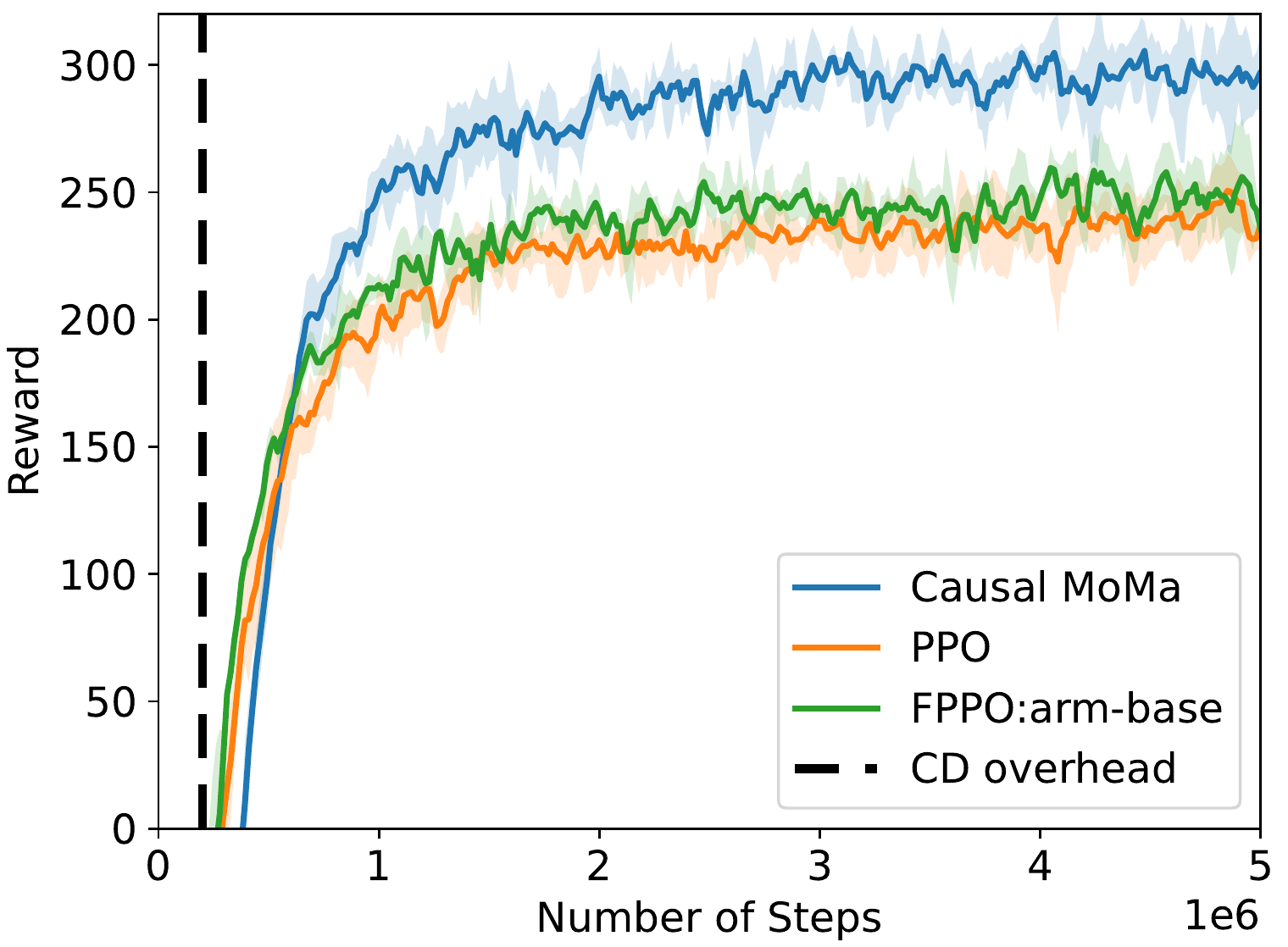}\hfill
\includegraphics[width=0.3\textwidth]{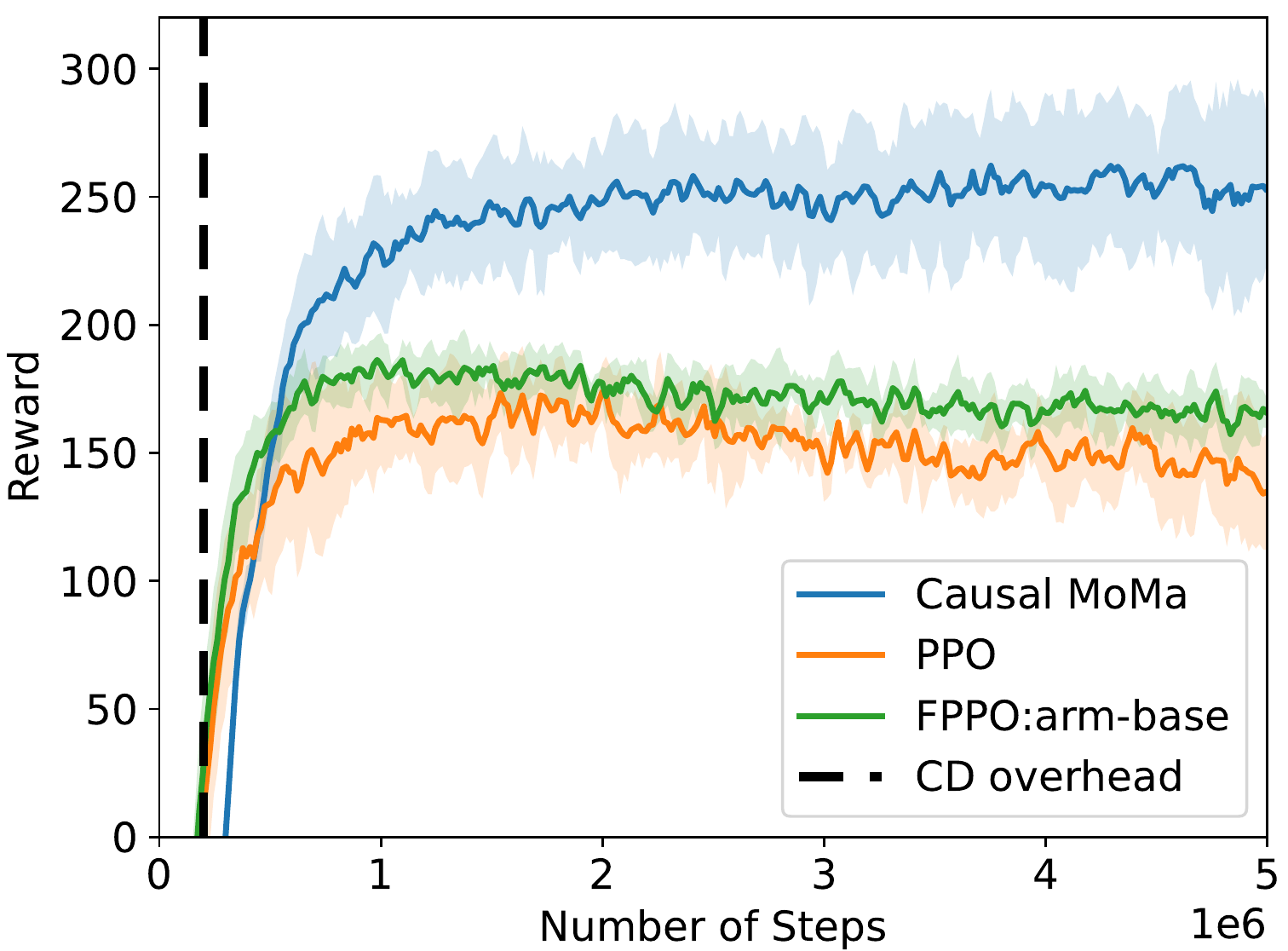}
\caption{Experimental evaluation of \methodname{} on the iGibson~\cite{li2022igibson} domain: (\textit{Left}) the agent is placed in one of eight possible household scenes and controls one of two realistically simulated mobile manipulation embodiments, a Fetch or an HSR robot, with continuous action dimensions and different dexterity (7 vs. 5 degrees of freedom in the arm, non-holonomic vs. holonomic base) for a virtual \textit{place glass} task: reaching a desired location with the hand while keeping a fixed hand orientation and avoiding collisions. Obstacles and robot initial locations are randomized per episode. (\textit{Middle and Right}) training curves for \methodname{} and baselines for Fetch (\textit{middle}) and HSR (\textit{right}) embodiments, five seeds each, mean and std. In this complex setup, \methodname{} consistently outperforms the baselines and achieves a higher return thanks to a reduced gradient variance with the causal policy gradient.} 
%the baselines converge to sub-optimal solutions due to large gradient variance in the non-factored action space.}
\label{fig:igibsonexp}
\end{figure*}

\section{Experimental Evaluation}
\label{s:exp}

We evaluate \methodname{} in three sets of MoMa environments: a simplistic simulated MoMa robot in Minigrid; realistic simulated MoMa robots in iGibson and Gazebo; and a physical MoMa robot in the real world. 
In the Minigrid simulator, the agent controls discrete actions, while in Gibson, Gazebo, and the real world, the action spaces are continuous.
We compare \methodname{} in simulation against three baseline reinforcement learning algorithms: ``vanilla'' PPO (not factored) and Factored PPO (FPPO) either with a randomly generated causal dependency or with a hardcoded arm-base separation and association to the reward terms based on domain knowledge (similar to the method by \citet{fu2022deep}). Additionally, in the Gazebo simulator, we compare against SLQ-MPC~\cite{pankert2020perceptive}, a reactive whole-body controller. In the real world, we compare against CBiRRT2~\cite{yamamoto2019development}, a modified version of the rapidly exploring random trees (RRT) sampling-based motion planner for whole-body motion planning combined with a trajectory executor. We compare against two baselines with CBiRRT2: an open-loop version (plan once and execute) and a replanning version that re-evaluates the path every \SI{3}{\second}. Both baselines have privileged access to the layout of the environment at the beginning of the episodes.

In our experiments, we aim at answering the following questions:
\textit{Q1:} Does the discovered causal matrix match the ground truth when the ground truth is available? (Sec.~\ref{ss:sim1})
\textit{Q2:} Does \methodname{} improve performance compared to baseline RL algorithms that do not make use of (or make use of the wrong) the causal structure between actions and rewards? (Sec.~\ref{ss:sim1}, Sec.~\ref{ss:rsim2})
\textit{Q3:} Is \methodname{} general enough to apply to different types of robots? (Sec.~\ref{ss:rsim2})
\textit{Q4:} Can the learned policy generalize to unseen environments and to a real robot? (Sec.~\ref{ss:rwexp})
\textit{Q5:} How do the policies trained with \methodname{} compare against sampling-based planners and reactive controllers? (Sec.~\ref{ss:rsim2}, Sec.~\ref{ss:rwexp})

In the following, we first introduce and analyze both sets of experiments in simulation, followed by a description and analysis of the experiments on the real-world mobile manipulator. Hyperparameters and network architectures can be found in Appendix~\ref{sec:minidetail}, \ref{app:ig}, and \ref{app:na}.

\subsection{Evaluation in the Minigrid Simulator}
\label{ss:sim1}

%--move \texttt{forward}, \texttt{backward}, \texttt{rotate\_left} and \texttt{rotate\_right}-- that transition the state to discrete values. 
Our first set of experiments is performed in the Minigrid~\cite{minigrid} environment (Fig.~\ref{fig:minigridexp}, left).
We build upon the \textit{Lava Gap} task in the original Minigrid, where a simulated agent (red triangle) has to navigate to a specified goal (green tile) while avoiding lava (orange tiles). 
We modify the original task by expanding the action space to 4 dimensions, with the first two action dimensions $a_{\text{up/down}}, a_{\text{left/right}}$ corresponding to navigation actions, and the other two action dimensions $a_{\text{arm1}}, a_{\text{arm2}}$ corresponding to virtual arm manipulation actions. Each of the action dimensions has three possible discrete values.
We also modify the scene so that virtual arm actions are necessary at different tiles (of different colors), creating a multi-objective MoMa task represented by a complex composite reward with five reward terms:
\begin{equation}
\begin{aligned} 
    R_{\text{minigrid}} &= R_{\text{up/down}}+ R_{\text{left/right}}+ R_{\text{org}}+ R_{\text{green}}+ R_{\text{blue}}
\end{aligned}
\end{equation}
% $R_{up/down}, R_{left/right}, R_{org}, R_{green}, R_{blue}$.

$R_{\text{up/down}}$ and $ R_{\text{left/right}}$ are decomposed navigation rewards that encourage the agent to move towards the goal from any tile. $R_{\text{org}}$ requires the agent to avoid stepping into the orange tiles (penalty). $R_{\text{green}}$ and $R_{\text{blue}}$ are virtual manipulation rewards that require the agent to perform a specific action using $a_{\text{arm1}}$ and $ a_{\text{arm2}}$ when in blue and green tiles with waves, respectively.
%These two types of tiles are not visible to the agent until it steps onto them; this way the causal dependency of $ R_{\text{green}}$ and $R_{\text{blue}}$ terms is only on virtual arm actions. 
A complete description of the mathematical definition of the reward terms can be found in Appendix~\ref{sec:minidetail}.
The observations of the agent are $6 \times 6$ images corresponding to the state of the environment with the location of the agent.
%, with the green and blue tiles masked out as empty tiles.
Our modified Minigrid domain is defined to represent MoMa tasks with a causal relation between action dimensions and reward terms that can be easily determined by a human so that we can evaluate the performance of \methodname{}'s causal discovery step (\textit{Q1}).

\begin{figure}[t]
\includegraphics[trim={0.5cm 0cm 0.5cm 0.5cm},clip,width=0.25\textwidth]{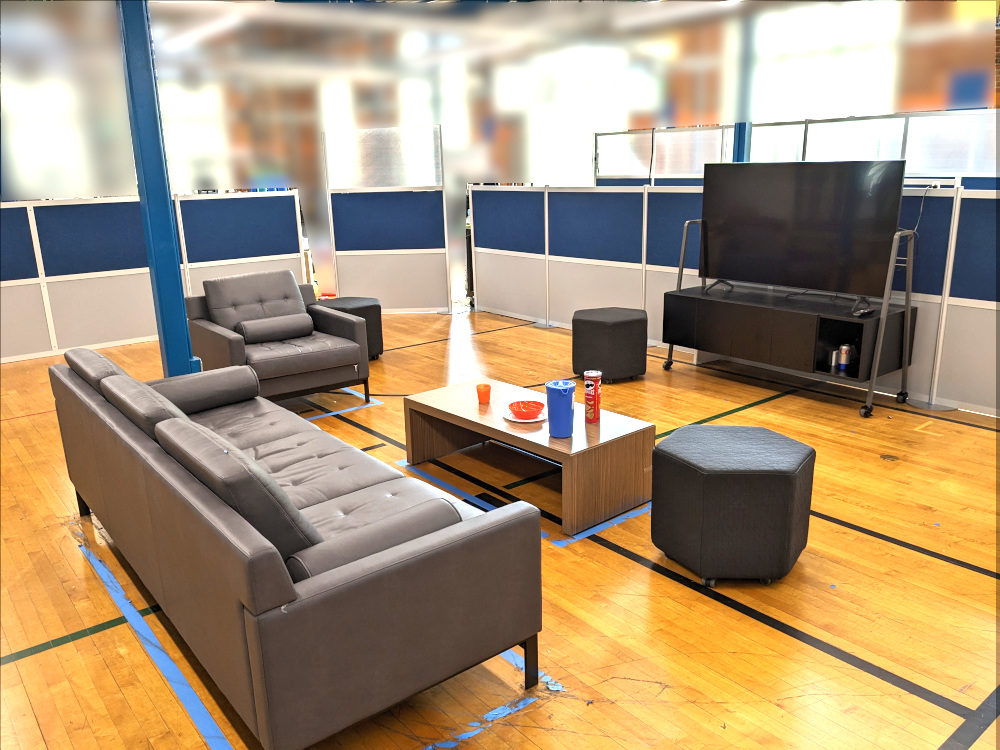}
\hfill
\includegraphics[trim={0cm 0cm 12cm 0},clip,width=0.23\textwidth]{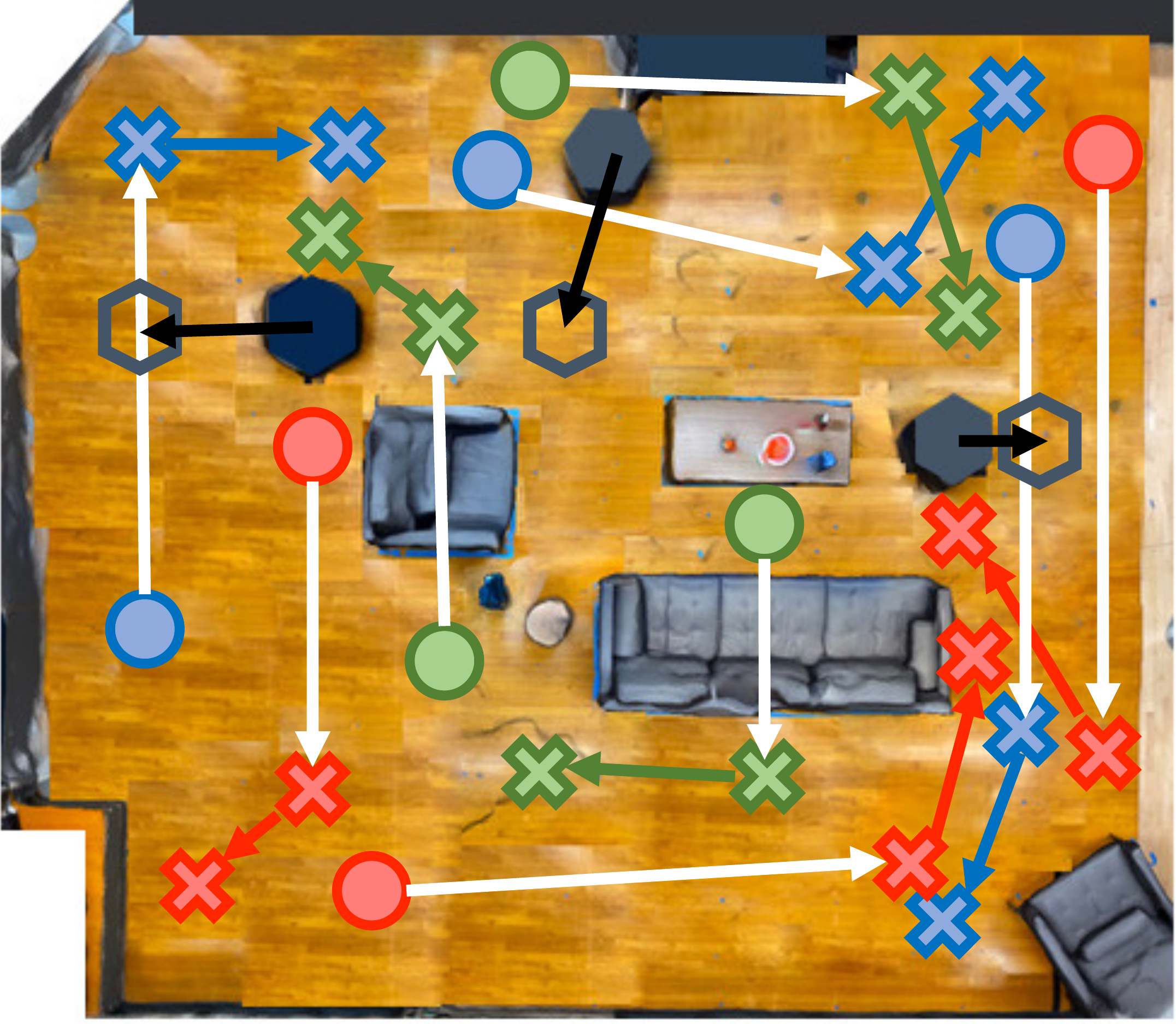}
\caption{Evaluation environment for \methodname{} in the real-world. \textit{Left:} The robot is placed in a mock apartment never seen during training and the best \methodname{} trained policy is transferred \textit{zero-shot}. \textit{Right:} the robot is tasked with reaching different locations with the end-effector (\textit{crosses}) from varying starting points (\textit{circles}) while keeping a desired orientation, avoiding collisions, and keeping the goal in sight. We evaluate paths with three types of obstacles, no obstacles (red), static obstacles (green), and dynamic obstacles (blue) in the direct path to the goal, frequent in household environments. Each setup repeats for two types of goals, static and dynamic. \textit{First cross}: robot's initial end-effector goal (and final for static goals); \textit{Second cross}: robot's final end-effector goal, when the goal is dynamic. The policy trained with \methodname{} achieves higher performance than a planning-based solution (with and without replanning) with privileged information about the layout of the scene.}
\label{fig:rwexp}
\end{figure}

\textbf{Results in Minigrid:} The ground truth causal matrix and the learned one discovered by \methodname{}, $B_{\text{Minigrid}}$, can be found in Appendix~\ref{sec:minidetail}.
%, together with the intermediate causal mutual-information (CMI) matrix\todo{do we talk about this matrix in the method section or appears only in the figure?} (see Fig.~\ref{fig:sysdiag}, top-right). 
We observe that the learned causal matrix matches exactly the ground truth, indicating the effectiveness of \methodname{}'s causal discovery step (\textit{Q1}).

Fig.~\ref{fig:minigridexp}, right, depicts the evolution of the reward during training for \methodname{} and baselines. 
Perhaps surprisingly, vanilla PPO fails in this seemingly simple domain, and converges to a local optimum of completely avoiding the colored tiles. This is because the agent is penalized by stepping onto the colored tiles when it does not execute the correct action in the virtual arm action dimension, and, at the early stages of training, most of the experiences lead to penalties. As a result, the vanilla PPO policy learns to avoid these tiles altogether, which prevents the agent from reaching the goal.
By comparison, \methodname{} avoids this local optimum by utilizing the discovered causal matrix to focus each action dimension on the reward terms that are causally related, and converges to a total return that significantly outperforms vanilla PPO (\textit{Q2}).
Interestingly, in the Minigrid domain, FPPO using a hardcoded arm-base separation and manual reward association is able to generate results that are almost as good as \methodname{}. 
This hardcoded causal dependency is included in Appendix~\ref{sec:minidetail}. Note that both arm dimensions are associated with both arm-related reward terms, an over-specification over the ground truth causal dependency.
These results indicate that, in simple domains, exploiting a slightly inaccurate causal dependency for causal policy gradient may still provide benefits over no-factorization, a possible reason for the common use of base/arm factorization in prior work. 
However, FPPO with a randomly generated causal dependency fails completely to train, achieving worse performance than non-factored PPO: a completely wrong factorization is catastrophic for the training process.

\subsection{Evaluation in Realistic Robot Simulators}
\label{ss:rsim2}
Our second set of experiments evaluates the performance of \methodname{} in two realistic robot simulators: iGibson 2.0~\cite{li2022igibson} and Gazebo~\cite{gazebo}. 
We carried out all the training in iGibson, a realistic simulation environment for mobile manipulators in household scenes (Fig.~\ref{fig:igibsonexp}, left). 
In iGibson, we experimented with two types of mobile manipulators, a Fetch and an HSR robot, both with continuous action spaces and modeled based on real-world robot platforms.
Fetch and HSR offer very different motion capabilities, which allows us to test whether our \methodname{} is general enough to be applied on different robots: Fetch is composed of a \textit{non-omnidirectional} base, a controllable pan-tilt head, and a \textit{7-DoF} robot arm controlled by moving the end-effector with Cartesian space commands. The HSR is composed of an \textit{omnidirectional} base, a controllable pan-tilt head, and an arm with only \textit{5-DoF} controlled with joint space commands. The agent controls the base with linear and angular velocity, where for Fetch the linear velocity is a 1-D scalar, and for HSR is a 2D vector. We compare against baselines and perform ablation studies in iGibson.
Finally, we tested zero-shot transferring the learned HSR policies in iGibson to the Gazebo simulator~\cite{gazebo} in order to compare against SLQ-MPC~\cite{pankert2020perceptive}, a reactive whole-body controller.

\begin{table*}[t]
    \caption{Real robot experiments on a set of tasks. Each entry in the table represents the success rate, averaged over nine runs (over three distinct layouts, three trials each). A run is considered successful if the robot can get to within $d$ distance of the target without collisions. Our method is able to zero-shot transfer to the real world and achieve better performance compared to a whole-body motion planner on five out of the six scenarios tested.}
    \label{tab:rw}
    \centering
    \resizebox{\textwidth}{!}{%
    %\begin{tabular}{ccrrrrrrrrrc}
    %  Threshold $T=$
    \begin{tabular}{cccccccccccc}
    \toprule
     Goal & Obstacles &  \multicolumn{2}{c}{Causal MoMa (ours)} & \multicolumn{2}{c}{CBiRRT2~\cite{yamamoto2019development}} & \multicolumn{2}{c}{CBiRRT2-replan~\cite{yamamoto2019development}} & \\
    \cmidrule(lr){3-4}\cmidrule(lr){5-6}\cmidrule(lr){7-8}
      & & d=\SI{0.15}{\meter} & d=\SI{0.3}{\meter} & d=\SI{0.15}{\meter} & d=\SI{0.3}{\meter} & d=\SI{0.15}{\meter} & d=\SI{0.3}{\meter} \\
     
     \midrule
     \multirow{3}{*}{Static Goal} 
        &No Obstacle & \textbf{9/9} & \textbf{9/9}& 6/9 & 8/9 & 8/9 & \textbf{9/9}  \\
        &Static Obstacles & 7/9 & 7/9 & 5/9 & \textbf{8/9} & \textbf{8/9} & \textbf{8/9}  \\
        &Dynamic Obstacles & \textbf{8/9} & \textbf{9/9}& 0/9 & 2/9 & 1/9 & 4/9  \\
    \midrule
    \multirow{3}{*}{Dynamic Goal} 
        &No Obstacle & \textbf{9/9 }& \textbf{9/9 }& 1/9 & 3/9 & 4/9 & 8/9  \\
        &Static Obstacles & \textbf{9/9 }& \textbf{9/9 }& 0/9 & 1/9 & 2/9 & 6/9 \\
        &Dynamic Obstacles & \textbf{8/9} & \textbf{9/9 }& 0/9 & 0/9 & 0/9 & 2/9 \\   
    \bottomrule
    \end{tabular}
    }
\end{table*}

The agent in the realistic simulator must achieve a MoMa task with multiple sub-objectives that are ubiquitous in household tasks: reaching a location with the end-effector and closing the hand, without collisions with the environment or self-collisions, while keeping the goal in the camera view and the hand in a predefined orientation and height. This corresponds to a composite reward with eight reward terms:
\begin{equation}
\begin{aligned} 
    R_{\text{iGibson}} &= R_{\text{reach}} + R_{\text{eef ori}} + R_{\text{eef height}} + R_{\text{base col}} \\& +  R_{\text{arm col}} +  R_{\text{self col}} +  R_{\text{head ori}} +  R_{\text{gripper}}
\end{aligned}
\end{equation}

$R_{\text{reach}}$ encourages the robot to reach a 3D goal with its end-effector (eef). It also contains a shaping component that rewards the robot every timestep if it gets closer to the goal. $R_{\text{eef ori}}$ encourages the robot to align its eef's orientation with a target orientation that is randomly sampled at the start of each episode. During deployment, the user can specify different target orientations for different purposes, e.g. holding a cup of water such that the water doesn't spill. $R_{\text{eef height}}$ specifies the desired eef height across the entire trajectory. $R_{\text{base col}}$, $R_{\text{arm col}}$, and $ R_{\text{self col}}$ are collision penalties for respective body parts of the robot. Notice that $R_{\text{base col}}$ and $R_{\text{arm col}}$ do not account for collisions between the base and the arm, which is managed by $ R_{\text{self col}}$. $R_{\text{head ori}}$ encourages the head-mounted camera to look in the direction of the goal, which helps the robot in the real world to maintain a good estimation of the relative goal position. $R_{\text{gripper}}$ encourages the robot to toggle the gripper when it is close to the goal. The HSR and Fetch experiments share the same reward function. The observation space for the HSR consists of a 25-dimensional proprioceptive and task-related observation vector and a 270-dimensional LiDAR scan. The observation space for the Fetch consists of a 27-dimensional proprioceptive and task-related observation vector and a 220-dimensional LiDAR scan. A complete description of the action and observation spaces and the mathematical definition of the reward terms can be found in Appendix~\ref{app:ig}.

\begin{figure*}[ht!]
\centering
\includegraphics[width=0.16\textwidth]{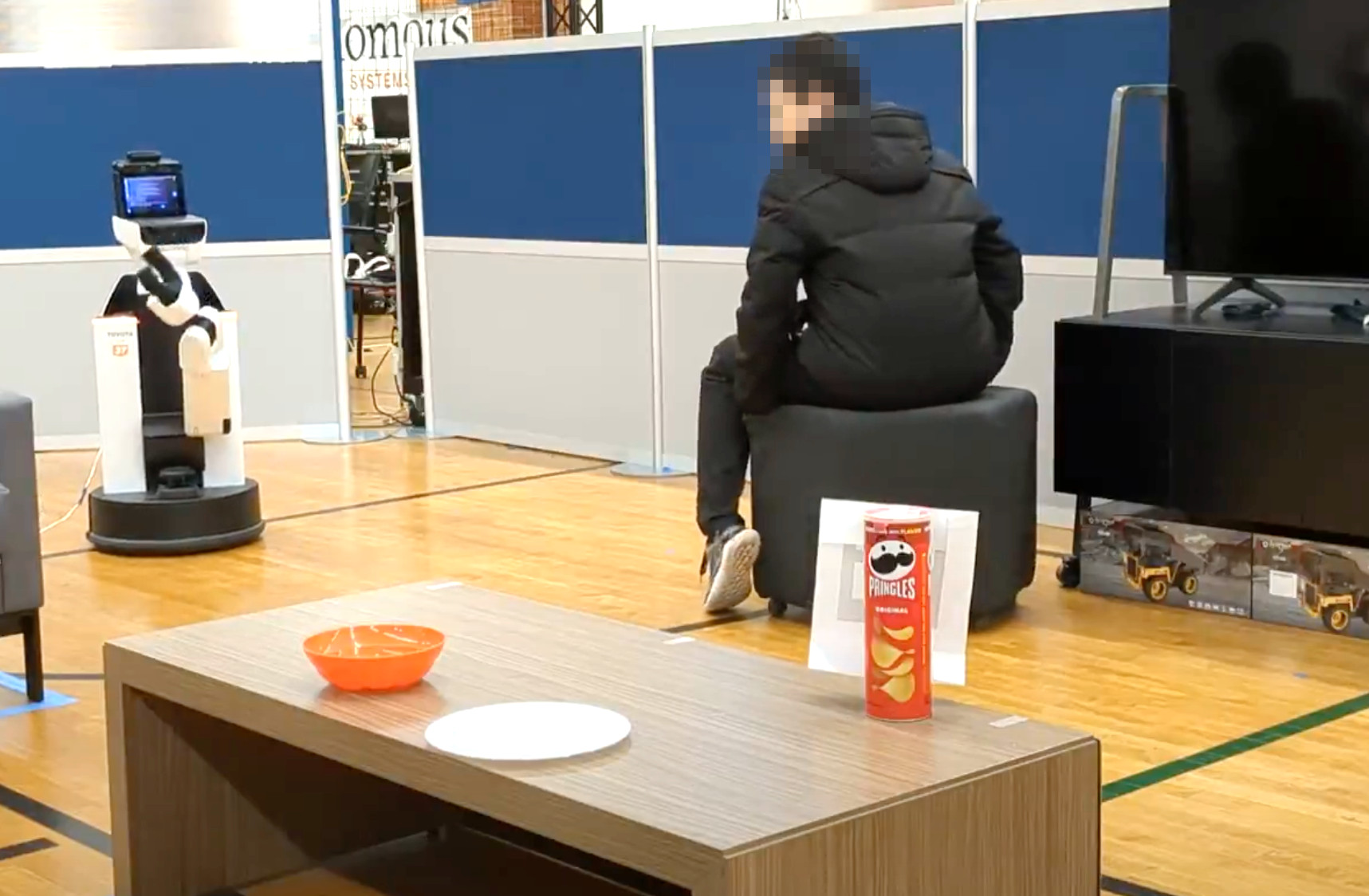}\hfill
\includegraphics[width=0.16\textwidth]{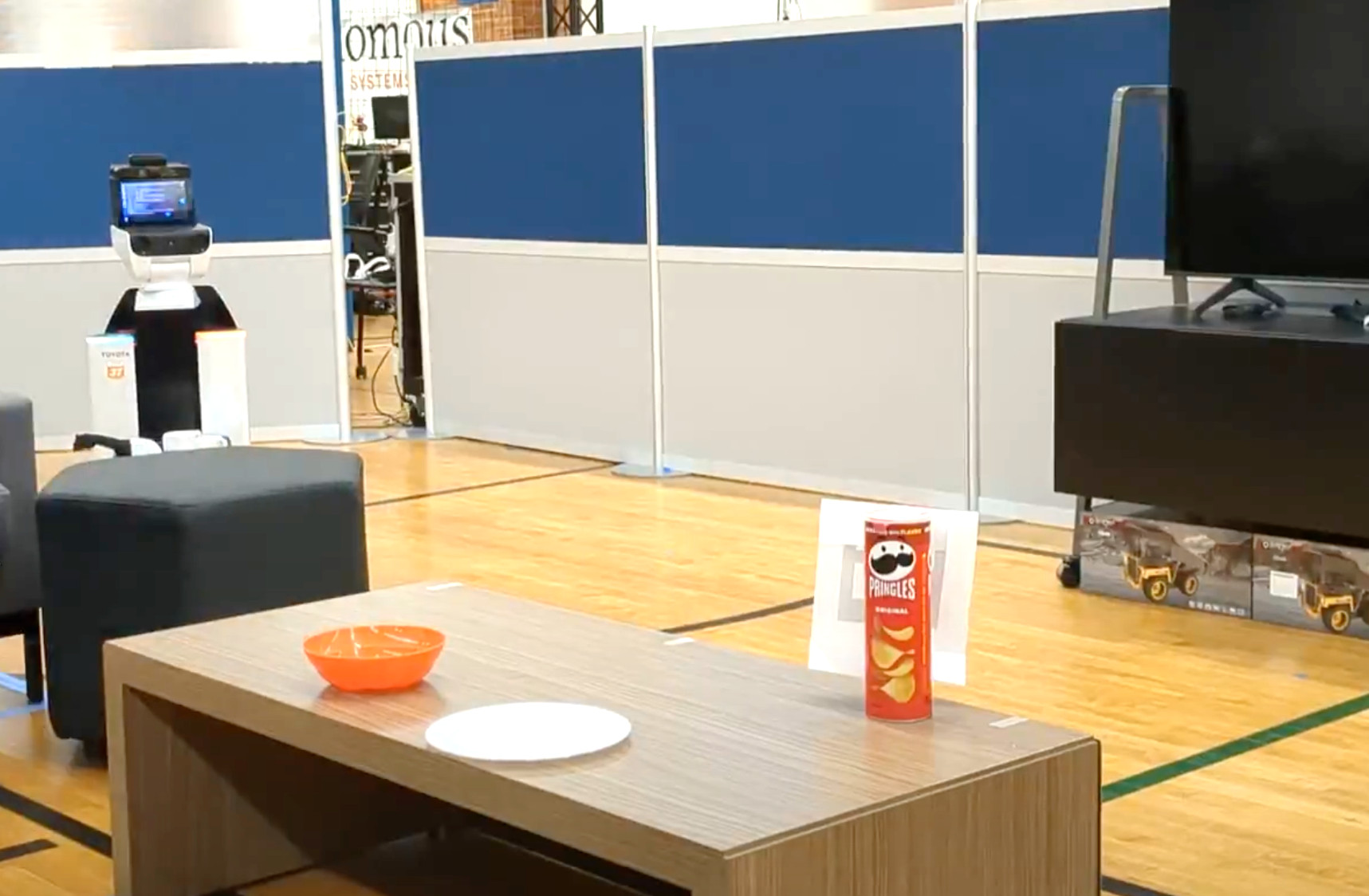}\hfill
\includegraphics[width=0.16\textwidth]{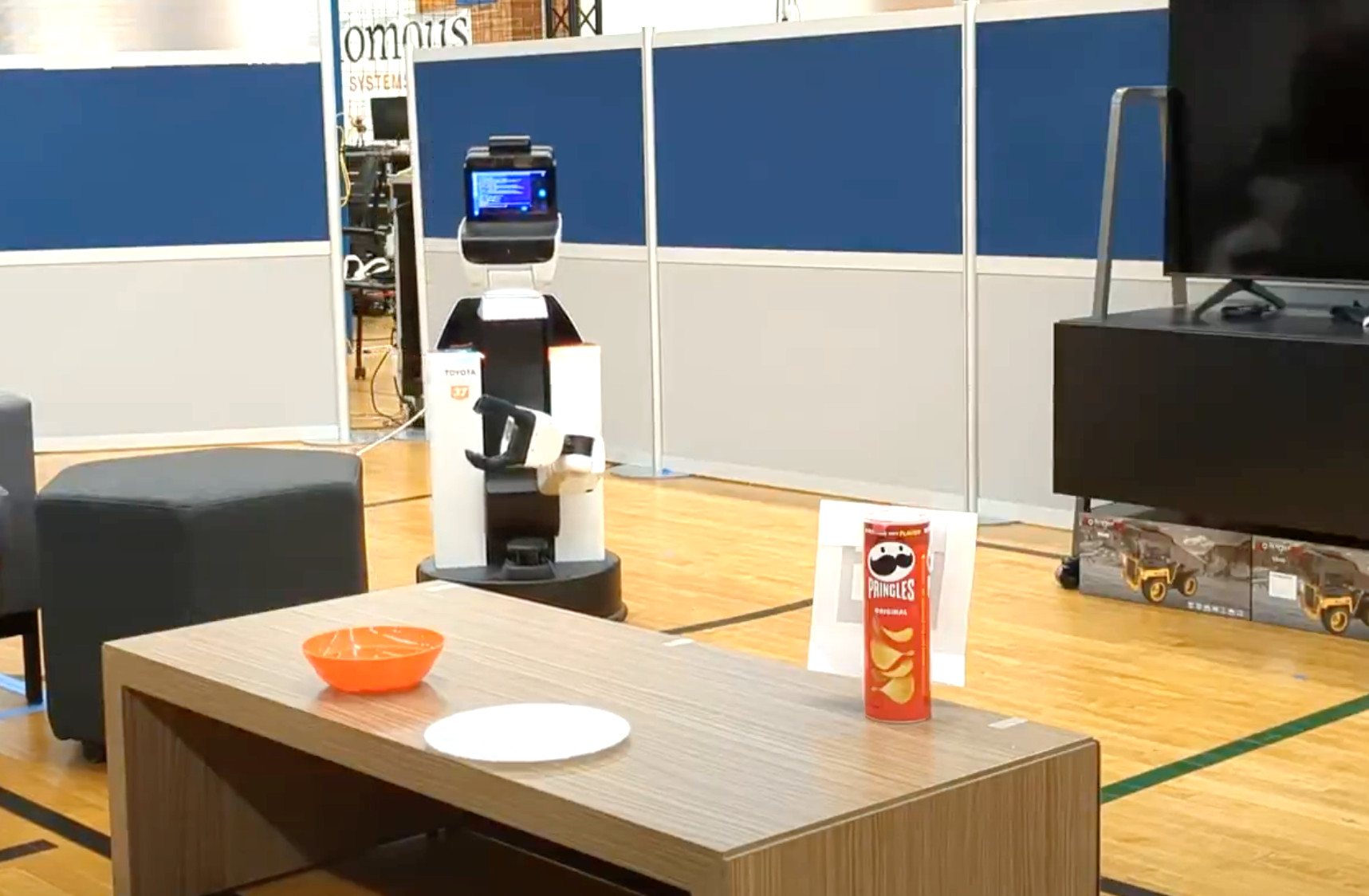}\hfill
\includegraphics[width=0.16\textwidth]{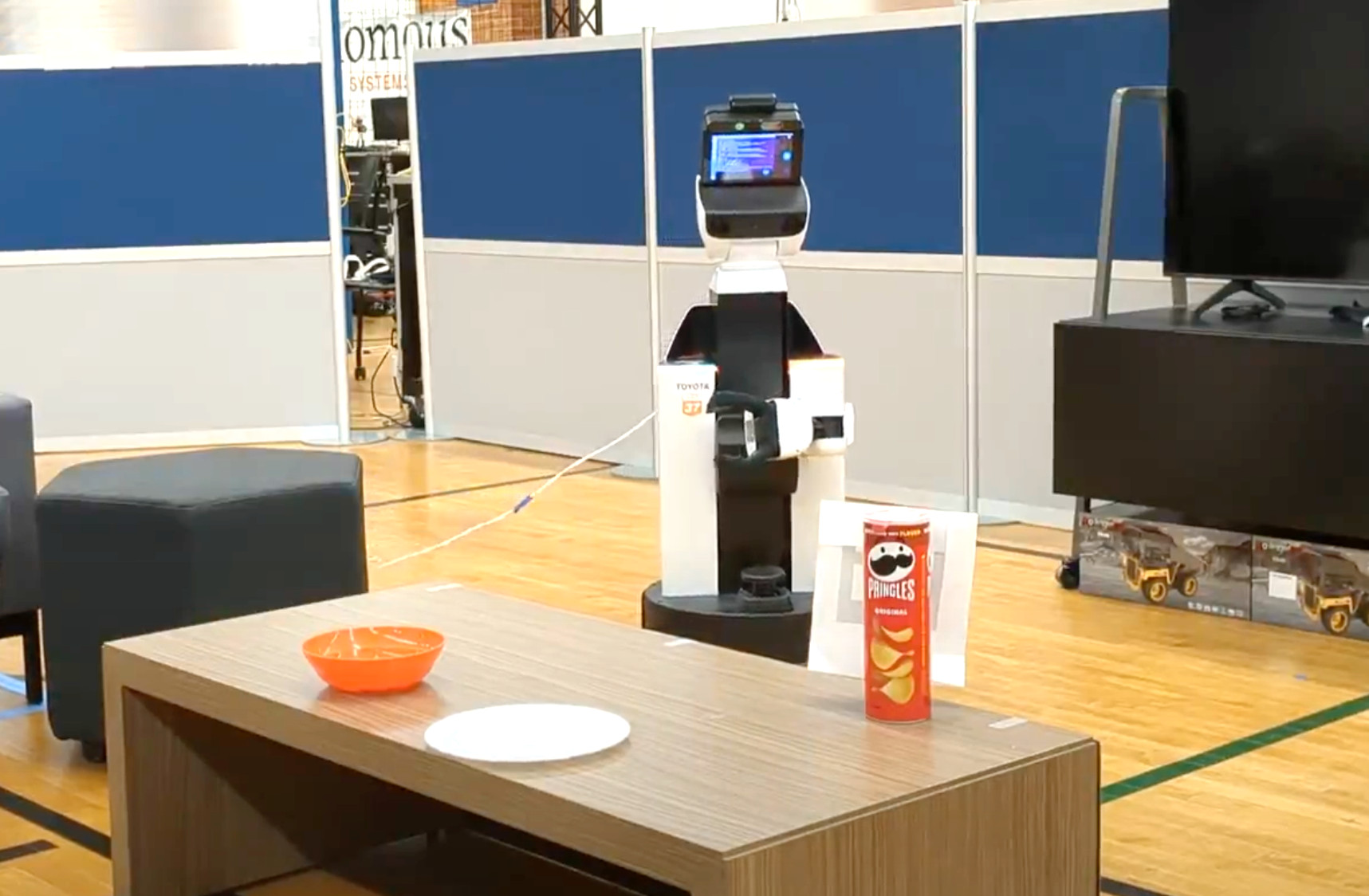}\hfill
\includegraphics[width=0.16\textwidth]{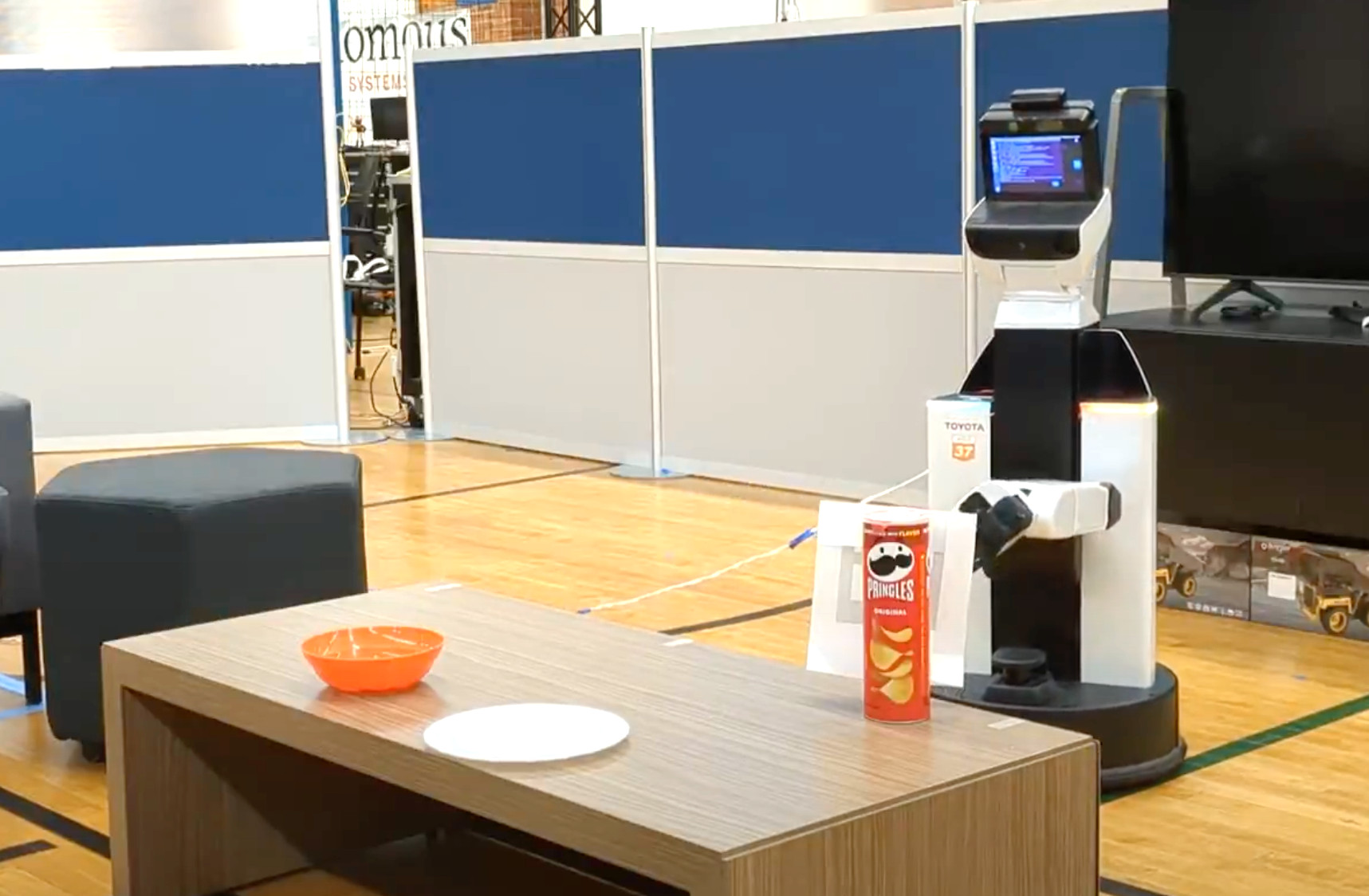}\hfill
\includegraphics[width=0.16\textwidth]{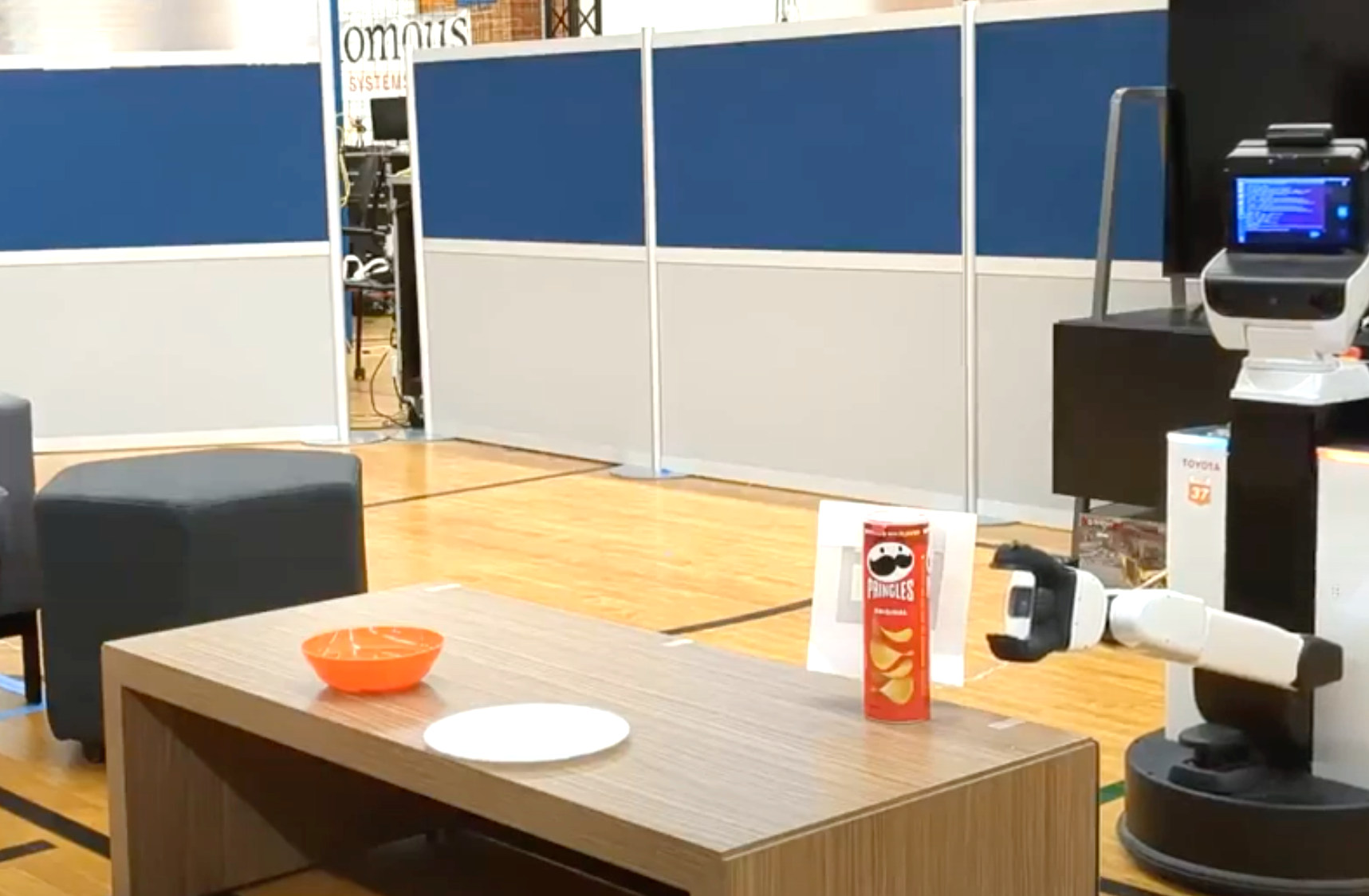}
\caption{Temporal evolution of one of our experiments with \methodname{} in the real world. The robot is tasked with reaching the object on the table marked with a tag while going around static and dynamic obstacles in the direct path, keeping the orientation of the end-effector upright, avoiding self-collisions, and keeping the target in view. The policy trained with \methodname{} can robustly achieve the task in this unseen environment, controlling simultaneously the whole body. More examples in the video attachment.}
\label{fig:realexample}
\end{figure*}

\textbf{Results in iGibson:} 
In iGibson, we obtain the main training results --causal matrices, reward curves, success rates-- of \methodname{}, RL baselines, and ablations.
The learned causal matrix for Fetch and HSR, $B_{\text{Fetch}}$ and $B_{\text{HSR}}$, can be found in Appendix~\ref{app:ig}.
Notice that $B_{\text{HSR}}$ is sparser than $B_{\text{Fetch}}$, suggesting that smaller subsets of the available action space dimensions are causally related to each reward term for HSR than Fetch.
% , and are thus necessary to optimize for them in the MoMa task. 
We hypothesize that this caused a larger improvement in the causal policy gradient in HSR than in Fetch and led to the larger differences in \methodname{} over the baselines observable in Fig.~\ref{fig:igibsonexp} (\textit{Q3}). 

Fig.~\ref{fig:igibsonexp}, depicts the evolution of the reward during training for \methodname{} and baselines for the iGibson experiment with Fetch (middle) and HSR (right). For both robot embodiments, \methodname{} achieves superior performance compared to both vanilla PPO and FPPO with arm-base separation, despite the initial computational overhead of performing causal inference (CD overhead - represented by the dotted line in the plots). The arm-base separation underperforms \methodname{}, suggesting that, for real-world problems with multiple objectives and high dimensional action spaces, more fine-grained causal relationships between action and reward terms are needed to fully take advantage of the factored nature of the problem (\textit{Q2}). 
Additionally, we examine the success rate of the different learning algorithms at the end of training in iGibson. 
A run is considered successful if the robot is able to reach within $d=\SI{0.15}{\meter}$ of the goal with its end-effector without collision. For each method on each robot type, we run 1000 trials with randomly generated start and goal positions. In these conditions, \methodname{} achieves 93.6\% and 92.1\% success on HSR and Fetch, significantly over Vanilla PPO (74.3\% HSR, 79.0\% Fetch) and FPPO with arm-base factorization (76.1\% HSR, 78.7\% Fetch).
% \vspace{2mm}
% \begin{center}
% \begin{tabular}{ c||c|c}

%  Method \textbackslash \hspace{0.1cm} Robot & HSR  & Fetch \\
%  \hline
%  \methodname{} (ours) & $\textbf{93.6}\%$   &$\textbf{92.1}\%$  \\
%  Vanilla PPO~\cite{schulman2017proximal} & $74.3\%$   &$79.0\%$  \\
%  FPPO arm-base~\cite{fu2022deep} & $76.1\%$   &$78.7\%$  \\

% \end{tabular}
% \end{center}
% \vspace{2mm}

% We can see that the higher reward obtained by \methodname{} indeed corresponds to a higher success rate compared to the baseline algorithms.

\textbf{Results in Gazebo}:
In Gazebo, we compare the policy trained with \methodname{} and the baselines against SLQ-MPC, a state-of-the-art model predictive control (MPC) strategy for reactive whole-body mobile manipulation. We use Gazebo as it is the platform where the implementation of SLQ-MPC is available; to that end, we zero-shot transfer our policies learned in iGibson into the Gazebo simulator (sim2sim) and evaluate them on four distinct scenarios based on the goal type (static/dynamic) and obstacle type (none/static/dynamic). When either the goal or the obstacle is dynamic, they move at \SI{0.3}{\meter/\second} along hand-crafted trajectories.
We report the success rates for the learned policies and SLQ-MPC, as well as additional descriptions of the Gazebo environment, in Appendix~\ref{app:eva}. 

We observe that the SLQ-MPC reactive approach improves over the baselines for the scenes with static goal and no obstacles or static ones, but it still underperforms compared to \methodname{}. The performance is significantly lower than \methodname{} in dynamic scenes because SLQ-MPC is unable to balance multiple objectives correctly and lacks an additional solution to map sensor signals to its internal model.
These failures reveal a fundamental limitation of model-based reactive control and planning methods: they require some additional mechanism to map sensor data to their internal model (e.g., defining, detecting, tracking obstacles, or creating and maintaining a map with SLAM\ldots), and struggle when such mechanisms are missing and the internal model becomes inaccurate. By contrast, \methodname{}'s policy does not require building and maintaining a geometric world model as it maps directly sensor signals to actions (\textit{Q5}).

\subsection{Evaluation on a Real-World Mobile Manipulator}
\label{ss:rwexp}
In a final set of experiments, we evaluate the performance of our policies trained in simulation with \methodname{} when transferred zero-shot to control a real robot, an HSR mobile manipulator, and compare against two baselines based on the sampling-based planner CBiRRT2~\cite{yamamoto2019development}, with and without replanning.
We use the published implementation of CBiRRT2 customized and tuned for the HSR robot.
% We evaluate the \methodname{} policy against CBiRRT2~\cite{yamamoto2019development}, a modified version of the rapidly exploring random trees (RRT) sampling-based motion planner for whole-body motion planning combined with a trajectory executor. We compare against two versions of CBiRRT2. 
In the first CBiRRT2-based baseline, an open-loop trajectory is computed at the beginning and is directly executed using a trajectory execution controller with spline interpolation and scheduled motion generation. The end of a CBiRRT trial is indicated by the robot terminating all the points in the trajectory and stopping.
The second baseline (CBiRRT2-replan) replans for a new trajectory every \SI{3}{\second} and executes with the trajectory controller. Both baselines have access to privileged information: a layout (obstacle map) of the environment; \methodname{} solely relies on onboard sensors.

We set up our experiment in a household scene (see Fig.~\ref{fig:rwexp}) that has never been seen by the agent during training. The robot is tasked with reaching a desired location with the end-effector while keeping the hand in a user-specified orientation and avoiding collisions with the environment and with itself. This MoMa task is one of the most general skills required for a mobile manipulator in household domains. \methodname{} and baselines track the relative goal location through a QR marker recognized with the robot's onboard head camera~\cite{yamamoto2019development}. Action and observation spaces are similar to the iGibson HSR setup.

We evaluate our method on six distinct scenarios based on the goal type (static/dynamic) and obstacle type (none/static/dynamic). 
Each scenario consists of three different initial and goal configurations, as depicted in Fig.~\ref{fig:rwexp}, right. 
When either the obstacle or the goal is dynamic, it moves at about \SI{0.3}{\meter/\second} along the trajectory indicated in Fig.~\ref{fig:rwexp}, right. We repeat each configuration three times to account for episodic randomness. Similar to Sec.~\ref{ss:rsim2}, we consider a run to be successful if the robot reaches within $d$ meters of the goal with the end-effector without collision. We evaluate results for two different $d$ values, corresponding to low precision and high precision conditions. 

\textbf{Results on the real mobile manipulator:} The results of our evaluation in the real world are summarized in Table~\ref{tab:rw}. Each row in the table corresponds to a different scenario (goal and obstacle type), and each entry is averaged over nine runs across three different layouts. Our method is able to outperform the baselines on all but one scenario (static goal with static obstacles) and has a significant advantage over the baselines in all the dynamic cases, indicating the capacity of our policy trained with \methodname{} to control the whole-body motion reactively (\textit{Q5}). Notice that our training environment does not include any dynamic obstacles, which indicates the robustness of our method and its generalization ability (\textit{Q4}). 

A typical failure mode for CBiRRT2 in the dynamic scenario is to either collide with a moving obstacle or lose track of the goal location, even with re-planning. 
Even in the easiest case (static goal, no obstacle), the open-loop execution of the CBiRRT2 plan is not able to reach the goal position very accurately due to the mismatch between the motion model of the robot and the real motion (possibly because of wheel slip or inaccuracies of the model), causing the executed plans to deviate from the planned ones. These results confirm our conclusions in Sec.~\ref{ss:rsim2} that model-based planning and reactive control methods struggle with dynamic problems when there is a lack of a perceptual system to update the internal model with observations. More advanced planning solutions integrating sensing~\cite{masehian2007robot,maly2013iterative,vidal2019online,aoude2013probabilistically} may alleviate this issue but would still be affected by the multiple objectives in the MoMa tasks.

Significantly, the lower success rate of \methodname{} under the static goal and static obstacles setting (second row) is caused by one particular set of initial and goal configurations, where the robot starts off in a local optimum blocked by a long obstacle, a sofa bed (green circle and cross at the center of Fig.~\ref{fig:rwexp}, right). In these conditions, \methodname{} is not able to reliably plan the long path required around the sofa based only on the partial observation from the LiDAR. By contrast, the CBiRRT2 planner with a privileged layout map can consider global information and plans a path around the obstacle, outperforming \methodname{} in this setup. 
Notice that in addition to reaching the goal, \methodname{} is also able to keep the robot's hand towards user-indicated orientations during the execution of the policy, as it was trained in simulation. We provide an analysis of the deviation of the orientation to the desired value over the course of an episode in Appendix~\ref{app:eef}. 

Finally, as a proof of concept, we demonstrate the potential of the skills learned with \methodname{} with two semantically meaningful MoMa tasks in the real robot: carrying a cup of water (beads) without spilling it and pouring it into a bowl, and following a person, while avoiding emerging obstacles. We demonstrate these behaviors in the supplementary video.

\section{Limitations and Conclusion} 
\label{sec:conclusion}

We introduced \methodname{}, a two-step procedure to train policies for Mobile Manipulation problems exploiting the causal dependencies between action space dimensions and reward terms. \methodname{} infers autonomously the underlying causal dependencies, removing the need for manually defining them, and exploits them in a causal policy gradient approach that leads to improved training performance through a reduction of the gradient variance. We demonstrated the benefits of training with \methodname{} for several tasks in simulation and the successful zero-shot transfer to an unseen real-world environment, where we control the whole body of a mobile manipulator around static and dynamic obstacles. 

However, \methodname{} is not without limitations. The most severe derive from possible violations of the assumptions in Sec.~\ref{s:method}. In particular, as indicated in assumption A4, \methodname{} cannot deal with very long causal dependencies between actions and rewards, e.g., in long-horizon tasks with sparse rewards. We hypothesize that a hierarchical causal discovery procedure at different horizon lengths would overcome this problem.
In addition, the policy generated with \methodname{} performs visuo-motor whole-body MoMa control and, as such, it may fail to plan long trajectories that avoid local minima. However, this problem was not deemed severe in our experiments.
Finally, our current policy does not consume images directly but transforms them into an end-effector goal position first; this is not a limitation of \methodname{} but rather a way to accelerate the training procedure. We are extending \methodname{} to include RGB observations. Even with these limitations, we hope \methodname{} simplifies the use of mobile manipulators in households and other environments.

\section*{Acknowledgments}
We thank the anonymous reviewers for their helpful comments on improving the paper. We thank members of RobIn and LARG for their valuable feedback on the idea formulation and manuscript. In particular, we thank Zizhao Wang for discussions on causal discovery, and Yuqian Jiang for discussions on real robot setup.

A portion of this work has taken place in the Learning Agents Research Group (LARG) at UT Austin.  LARG research is supported in part by NSF (CPS-1739964, IIS-1724157, FAIN-2019844), ONR (N00014-18-2243), ARO
(W911NF-19-2-0333), DARPA, Bosch, and UT Austin's Good Systems grand
challenge.  Peter Stone serves as the Executive Director of Sony AI
America and receives financial compensation for this work.  The terms of
this arrangement have been reviewed and approved by the University of
Texas at Austin in accordance with its policy on objectivity in
research.
%% Use plainnat to work nicely with natbib. 

\bibliographystyle{plainnat}
\bibliography{references}

% \todo{Remove the appendix when submitting}
\clearpage
\appendix{}
\balance

\subsection{Proof of Theorem~\ref{thm} (Causal Sufficiency and Necessity)}
\label{app:mc}
\setcounter{section}{1}
As a first step to prove Theorem~\ref{thm}, we prove the following lemma, which is a local version of Theorem~\ref{thm}:
\begin{lemma}
\label{lemma:local}
$\forall i,j: $ $a_i^{t:t+k} \rightarrow r_j^{\sum{t:t+k}}$ if and only if $a_i^{t:t+k} {\not\!\perp\!\!\!\perp}  r_j^{\sum{t:t+k}} | \{s^t, \mathbf{a}^{t:t+k} \backslash a_i^{t:t+k}\}$
\end{lemma}

First, we consider the forward direction and prove that that: $$ a_i^{t:t+k} \not\!\perp\!\!\!\perp  r_j^{\sum{t:t+k}} | \{s^t, a^{t:t+k} \backslash a_i^{t:t+k}\} \implies a_i^{t:t+k} \rightarrow r_j^{\sum{t:t+k}} $$ Base on A3 (action independence), $a^{t:t+k}$ have no parent in the causal graph. Therefore, there cannot exist a confounder $Q$ with confounding path at $a_i^{t:t+k} \leftarrow Q \rightarrow r_j^{\sum{t:t+k}}$. 

Next, following A1 (causal Markov assumption) and A2 (causal faithfulness), conditioning on all other potential parents of $r_j^{\sum{t:t+k}}$ (i.e., $\{s^t, a^{t:t+k} \backslash a_i^{t:t+k}\}$), the conditional dependence at $a_i^{t:t+k} \not\!\perp\!\!\!\perp  r_j^{\sum{t:t+k}} | \{s^t, a^{t:t+k} \backslash a_i^{t:t+k}\}$ must result from the existence of the edge at $a_i^{t:t+k} \rightarrow r_j^{\sum{t:t+k}} $. 

Next, we prove the converse, i.e., $$a_i^{t:t+k} \rightarrow r_j^{\sum{t:t+k}} \implies a_i^{t:t+k} { \not\!\perp\!\!\!\perp}  r_j^{\sum{t:t+k}} | \{s^t, a^{t:t+k} \backslash a_i^{t:t+k}\}$$
Notice that this directly follows from the causal faithfulness assumption. This completes the proof for Lemma~\ref{lemma:local}.

Finally, assuming A4, we have $ a_i \rightarrow r_j $ if and only if $a_i^{t:t+k} \rightarrow r_j^{\sum{t:t+k}}$  for some $t$. This means that $a_i \rightarrow r_j$,  $a_i^{t:t+k} \rightarrow r_j^{\sum{t:t+k}}$, and $a_i^{t:t+k} {\not\!\perp\!\!\!\perp}  r_j^{\sum{t:t+k}} | \{s^t, \mathbf{a}^{t:t+k} \backslash a_i^{t:t+k}\}$ are materially equivalent. This completes the proof. $\blacksquare$

\subsection{Proof Sketch of Theorem~\ref{thm_fpg} (Causal Policy Gradient)}
\label{ss:prof_cpg}
\setcounter{section}{2}
First, we summarize the Factored Policy Gradient (FPG) proposition derived in \citet{spooner2021factored}:

\begin{prop}
\label{prop:fpg}
Take a $\Sigma$-factored policy $\pi_\theta(\mathbf{a}|s)$, a causal matrix for the  $\Sigma$-factored action space $K_\Sigma$, and $|\theta| \times |\Sigma|$ matrix of scores $S(s, \mathbf{a})$. Then, for target vector $\Psi(s, \mathbf{a})$ and multipliers $\lambda$, the FPG estimator
$g^F(s, \mathbf{a}) = S(s, \mathbf{a}) K_\Sigma \lambda \circ \Psi(s, \mathbf{a})$ is an unbiased estimator of the true policy gradient.
\end{prop}
We refer the reader to \citet{spooner2021factored} for detailed proof and terminology definition of this proposition. 

Now, we set the policy factorization $\Sigma$ to be per-action-dimension factorization, which turn $S(s, \mathbf{a})$ to $\nabla_{\theta} \log \pi_{\theta} (\mathbf{a} | s)$ and $K_\Sigma$ to $B_{\text{true}}$. Next, we use advantage $\hat{\mathbb{A}}^\pi(s, \mathbf{a})$ in place of the target vector $\Psi(s, \mathbf{a})$, and set multipliers $\lambda$ (which represents the weight of each reward term) to be $\mathbf{1}$, to obtain the modified FPG estimator $g^F_{mod}(s, \textbf{a}) = \nabla_{\theta} \log \pi_{\theta} (\textbf{a} | s) \cdot {B_{\text{true}}} \cdot \hat{\mathbb{A}}^\pi(s, \mathbf{a}) $ 

The derivation above indicates that Eq.~\ref{eqn:fpg} used in \methodname{} is a special case of the Factored Policy Gradient, and thus it follows that our Eq.~\ref{eqn:fpg} is an unbiased estimator of the true policy gradient. $\blacksquare$

\subsection{Minigrid Experimental Details}
\label{sec:minidetail}
% We explain the detail of the RL experiments in simulation in depth, including the observation space, exact reward functions, action space, and the causal matrix discovered by our method.

% \subsubsection{}
\textbf{Reward Function:} The reward function in the Minigrid domain is defined by
\begin{equation}
\begin{aligned} 
    R_{\text{minigrid}} &= R_{\text{up/down}}+ R_{\text{left/right}}+ R_{\text{org}}+ R_{\text{green}}+ R_{\text{blue}}
\end{aligned}
\end{equation}

$\bullet$ $R_{\text{up/down}} $ and $R_{\text{left/right}}$ rewards with +1 every time the agent moves closer to the goal in the vertical/horizontal direction, and with -1 if it moves further away.

$\bullet$ $R_{\text{org}}$ penalizes with -5 every time the agent steps onto an orange tile.

$\bullet$ $R_{\text{green}}$ and $R_{\text{blue}}$ penalizes with -5 every time the agent steps out from a green/blue tile with waves without executing the correct arm1 or arm2 action, where the correct arm1/arm2 action is defined by the number of empty tiles (black) around the tile location.

\textbf{Observations:} The observations in the Minigrid domain are $6 \times 6$ images. These images have two channels: one channel that indicates the agent's location, and another channel for the layout of the grid, where each pixel corresponds to a tile and can take five different integer values indicating the five different types of tiles: blue tile with waves, green tile with waves, orange tile with waves, goal (green, no waves), empty (black).

\textbf{Causal Matrices in Minigrid:} Given the reward explained above, we can derive the \textit{ground truth} causal dependency between action space dimensions and reward terms:

\renewcommand{\kbldelim}{(}% Left delimiter
\renewcommand{\kbrdelim}{)}% Right delimiter
\[
  B^{\text{gt}}_{\text{minigrid}} = \kbordermatrix{
    & R_{\text{up/down}} & R_{\text{left/right}} & R_{\text{org}} & R_{\text{green}} & R_{\text{blue}} \\
    a_{\text{up/down}} & 1 & 0 & 1 & 0 & 0 \\
    a_{\text{left/right}} & 0 & 1 & 1 & 0 & 0 \\
    a_{\text{arm1}} & 0 & 0 & 0 & 1 & 0 \\
    a_{\text{arm2}} & 0 & 0 & 0 & 0 & 1 
  }
\]
where the decomposed locomotion reward ($R_{\text{up/down}} \& R_{\text{left/right}}$) are causally related to the respective locomotion action ($a_{\text{up/down}} \& a_{\text{left/right}}$), the locomotion penalty ($R_{\text{org}}$) are related to both of the locomotion actions, and the manipulator rewards ($R_{\text{green}} \& R_{\text{blue}}$) are only causally related to the respective manipulation action ($a_{\text{arm1}} \& a_{\text{arm2}}$).

The learned causal matrix, $B^{\text{cmi}}_{\text{minigrid}}$, using $k=1$ presents the following form:

\renewcommand{\kbldelim}{(}% Left delimiter
\renewcommand{\kbrdelim}{)}% Right delimiter
\[
  B^{\text{cmi}}_{\text{minigrid}} = \kbordermatrix{
    & R_{\text{up/down}} & R_{\text{left/right}} & R_{\text{org}} & R_{\text{green}} & R_{\text{blue}} \\
    a_{\text{up/down}} & 1 & 0 & 1 & 0 & 0 \\
    a_{\text{left/right}} & 0 & 1 & 1 & 0 & 0 \\
    a_{\text{arm1}} & 0 & 0 & 0 & 1 & 0 \\
    a_{\text{arm2}} & 0 & 0 & 0 & 0 & 1 
  }
\]
which corresponds exactly to the ground truth causal matrix.

In Minigrid, for our experiments with predefined arm-base separation, we group together the locomotion actions and the manipulation actions and map them manually in the most optimal manner to the different reward terms. This leads to the following causal matrix as best association between groups of action dimensions and reward terms:

\renewcommand{\kbldelim}{(}% Left delimiter
\renewcommand{\kbrdelim}{)}% Right delimiter
\[
  B^{\text{arm-base}}_{\text{minigrid}} = \kbordermatrix{
    & R_{\text{up/down}} & R_{\text{left/right}} & R_{\text{org}} & R_{\text{green}} & R_{\text{blue}} \\
    a_{\text{up/down}} & 1 & 1 & 1 & 0 & 0 \\
    a_{\text{left/right}} & 1 & 1 & 1 & 0 & 0 \\
    a_{\text{arm1}} & 0 & 0 & 0 & 1 & 1 \\
    a_{\text{arm2}} & 0 & 0 & 0 & 1 & 1 
  }
\]
This matrix is an over-specification of the causal dependency: all action dimensions that are causally related to each reward term are correctly indicated (no false-negatives) but there are some action dimensions that are additionally indicated and should not (some false-positives). In these conditions, the second step of \methodname{}, training the policy with causal policy gradient, can still leverage some of the benefits of the factorization but suffers from additional variance in the gradient estimation compared to the learned causal matrix, leading to the slightly worse performance observed in our experiments (Fig.~\ref{fig:minigridexp}, right).

\textbf{Hyperparameters:} The hyperparameters for the Minigrid experiments are included in the table below:

\begin{center}
\begin{tabular}{|c | c|} 
 % \hline
 % Hyperparameter Name  & Value \\ [0.5ex] 
 % \hline\hline
 \multicolumn{2}{c}{Causal Discovery Parameters} \\
 \hline
 CMI threshold & 0.02 \\ 
 \hline
 k (time interval) & 1 \\
 \hline
 Learning rate & 1e-4 \\
 \hline
 Batch size & 64 \\
 \hline
 Gradient clipping norm & 20 \\
 \hline
 \multicolumn{2}{c}{PPO Parameters} \\
 \hline
 Learning rate & 1e-3 \\
 \hline
 GAE $\lambda$ & 0.95 \\
 \hline
  Discount factor $\gamma$ & 0.99 \\
 \hline
  target-KL & None \\
\hline
  PPO clip range & 0.2 \\
 \hline
 Advantage Normalization & False \\
 \hline
\end{tabular}
\end{center}

For all PPO hyperparameters that are not specified in the table, we use the default value in stable baselines3~\cite{raffin2019stable}.

\subsection{iGibson Experiments Details}
\label{app:ig}
\textbf{Reward Function:} For each timestep, $t$, we denote the 3D position of the robot end-effector as $\text{pos}^t_{\text{eef}}$, the orientation of the robot end-effector with respect to base link as $ \text{ori}^t_{\text{eef}}$, the 3D position of the goal as $\text{pos}_{\text{goal}}$, and the target local orientation of the end-effector as $ \text{ori}^t_{\text{goal}}$. We use $s^t$ to denote the state vector that encompasses all the variables above. We use $d$ to denote distance, which is the L2 distance for positions and the arc distance for orientations.

The reward function for iGibson is defined by
\begin{equation}
\begin{aligned} 
    R_{\text{iGibson}} &= R_{\text{reach}} + R_{\text{eef ori}} + R_{\text{eef height}} + R_{\text{base col}} \\& +  R_{\text{arm col}} +  R_{\text{self col}} +  R_{\text{head ori}} +  R_{\text{gripper}}
\end{aligned}
\end{equation}

$\bullet$ $ R_{\text{reach}}$ = $ R_{\text{reach}}^{\text{pot}} + R_{\text{reach}}^{\text{point}}$, where:\\ 
\begin{equation}
\begin{aligned}
&R_{\text{reach}}^{\text{point}}(s^t)=
\begin{cases}
    20,& \text{if } d (\text{pos}_{\text{goal}}, \text{pos}^t_{\text{eef}})\leq 0.1\\
    0,              & \text{otherwise}
\end{cases} \\
&R_{\text{reach}}^{\text{pot}}(s^t) = d (\text{pos}_{\text{goal}}, \text{pos}^{t-1}_{\text{eef}}) - d (\text{pos}_{\text{goal}}, \text{pos}^t_{\text{eef}}) 
\end{aligned}
\end{equation}

$\bullet$ $R_{\text{eef ori}}(s^t) = -d( \text{ori}^t_{\text{goal}},  \text{ori}^t_{\text{eef}}) * 0.25$ penalizes deviating from a desired height.

$\bullet$ $R_{\text{eef height}}(s^t) = -d (\text{pos}_{\text{goal}}.z, \text{pos}^t_{\text{eef}}.z)* 0.25$ penalizes deviating from a desired orientation. During training, we use this reward combined with an orientation goal that is uniformly sampled and kept constant during each episode. In our real robot experiments, we demonstrate the generalization capabilities of the robot to change orientation by setting two different values depending on the distance between the end-effector and the position goal, $\text{pos}_{\text{goal}}$, one value when the distance is larger than \SI{1}{\meter} and a different one for closer distances. This emulates real-world tasks such as pouring a substance into a container placed at the final position, as illustrated in Fig.~\ref{fig:figure1}.

$\bullet$ $R_{\text{base col}},  R_{\text{arm col}}, \text{and} R_{\text{self col}}$ all incur a penalty of $-1$ if a corresponding collision happens at a given time step.

$\bullet$ $R_{\text{head ori}}$ gives a reward of 0.2 at every time step if the goal is within the field of view of the camera, and -0.2 otherwise.

$\bullet$ $R_{\text{gripper}}$ requires that the robot close the gripper when $d (\text{pos}_{\text{goal}}, \text{pos}^t_{\text{eef}})\leq 1 $, and open otherwise. If not, a cost of -0.2 will be incurred at each timestep.

\textbf{Task specification:} At the start of each episode, we randomly sample a 3D goal location, a robot starting location, and a target local end-effector orientation. Every time the robot reaches a goal, a new goal is automatically generated. Therefore, the robot is trying to reach as many goals as possible within an episode. The length of each episode is set to be 500. 

\textbf{Observations:} Both Fetch and HSR agents are provided an observation composed of a LiDAR reading, proprioceptive information, and a task observation vector.
The LiDAR reading of Fetch and HSR are both down-sampled to an angular resolution of $1^\circ$, resulting in 220-dimensional and 270-dimensional readings respectively.
The proprioceptive information for both the Fetch and the HSR consists of the robot's base velocity, the end-effector's current local position, the end-effector's current local orientation, joint positions, and the current timestep.
The task observation vector for both the Fetch and the HSR consists of the robot's relative distance to the goal, and the end-effector's goal orientation. All orientations are represented as quaternions.

\textbf{Action Space:} We used different action spaces for the HSR and Fetch robot. For Fetch, we use Cartesian space control for the arm, resulting in an 11-dimensional action space consisting of 2D locomotion actions for linear and angular velocity with the non-omnidirectional base ($a_{\text{forward}}, a_{\text{turn}}$), 2D head actions ($a_{\text{pan}}, a_{\text{tilt}}$), 
3D arm position actions ($a_{\text{arm.x}}, a_{\text{arm.y}}, a_{\text{arm.z}}$), 
3D arm orientation actions ($a_{\text{arm.rx}}, a_{\text{arm.ry}}, a_{\text{arm.rz}}$), 
and 1D gripper action ($a_{\text{grip}}$).
The arm Cartesian space actions are delta motion with respect to the current end-effector pose, they are defined with respect to the base link of the robot and converted to joint-level actions using an inverse-kinematics solver.
For HSR, we directly operate in joint space. The action space is also 11-dimensional, consisting of 3D locomotion actions for linear and angular velocity with the omnidirectional base ($a_{\text{forward}}, a_{\text{side}}, a_{\text{turn}}$), 2D head actions ($a_{\text{pan}}, a_{\text{tilt}}$), 
5D manipulation joint actions ($a_{\text{arm.lift}}, a_{\text{arm.flex}}, a_{\text{arm.roll}}, a_{\text{wrist.flex}}, a_{\text{wrist.roll}}$),  
and 1D gripper action ($a_{\text{grip}}$). Joint space actions are deltas with respect to the current joint configuration.

\textbf{Causal Matrices in iGibson:} For the Fetch robot, the causal matrices discovered by \methodname{} are presented in Fig.~\ref{fig:fetchcausal}. The HSR causal matrices discovered by \methodname{} are presented in Fig.~\ref{fig:hsrcausal}. In these experiments, we are not able to define ground truth causal matrices but from a reasoned comparison between the learned matrices and the reward terms (and from the benefits observed by using them through causal policy gradient in Fig.~\ref{fig:igibsonexp}) we conclude they are close to the underlying causal dependency. 

We also include in Fig.~\ref{fig:hsrcausal} and Fig.~\ref{fig:fetchcausal} the result of using a classical separation into base action dimensions and arm action dimensions and associating each group to the most related reward terms. In our experiments, this classical hardcoded factorization failed to train successful policies. We believe this is the case because they do not reflect all action dimensions that should be causally related to each reward term (some false negatives), making it impossible to learn to correct those actions through reinforcement.

\textbf{Hyperparameters:} We use the same hyperparameters for the experiments with Fetch and HSR robots indicating some versatility of our solution (no need to fine tuning per robot). The parameters are summarized in the table below:

\begin{center}
\begin{tabular}{|c | c|} 
 % \hline
 % Hyperparameter Name  & Value \\ [0.5ex] 
 % \hline\hline
 \multicolumn{2}{c}{Causal Discovery Parameters} \\
 \hline
 CMI threshold & 0.02 \\ 
 \hline
 k (time interval) & 1 \\
 \hline
 Learning rate & 1e-4 \\
 \hline
 Batch size & 64 \\
 \hline
 Gradient clipping norm & 20 \\
 \hline
 \multicolumn{2}{c}{PPO Parameters} \\
 \hline
 Learning rate & 5e-5 \\
 \hline
 GAE $\lambda$ & 0.95 \\
 \hline
  Discount factor $\gamma$ & 0.99 \\
 \hline
  target-KL & 0.15 \\
\hline
  PPO clip range & 0.2 \\
 \hline
 Advantage Normalization & True \\
 \hline
\end{tabular}
\end{center}

For all PPO hyperparameters that are not specified in the table, we use the default value in stable baselines3~\cite{raffin2019stable}.

\subsection{Network Architectures}
\label{app:na}
We denote convolution layers as $C(n,k,s)$, with $n$ being the number of kernels, $k$ being the kernel size, and $s$ being the stride; fully connected layer as $F(n)$, with $n$ being the output size; max pooling layer as $M(n)$, with $n$ being the pooling size; Flattening as $L$.

\textbf{State Feature Extractor Networks:}
For both policy learning and causal discovery, states are pre-processed by a state feature extractor network to obtain state features vector. 
Notice that the state feature extractor network for policy learning and causal discovery share the same architecture but not the same weight.
The state feature extractor network is shown below, with all $C$ and $F$  followed by a ReLU activation function except for the output layer:\\
$\bullet$ Minigrid: $C(16, 2, 1) - M(2) - C(32, 2, 1) - C(64, 2, 1) - L$\\
$\bullet$ iGibson: the LiDAR scan is passed through $C(32, 8, 4) - C(64, 4, 2) - C(64, 3, 1) - L$; the task observation vector is passed through $R(128)$. The two resulting vectors are then concatenated.

\textbf{Causal Discovery Networks:}
iGibson and Minigrid share the same causal discovery network architecture,  with all $F$  followed by a ReLU activation function except for the output layer:\\ 
$\bullet$ Feature Mapping $f()$: $F(128) - F(128)$\\
$\bullet$ Prediction Network $g()$ : $F(128) - F(128) - F(1)$\\

\textbf{Policy Learning Networks:}
All $F$ in policy learning are followed by a Tanh activation function except for the output layer:\\
$\bullet$ Minigrid policy network: $F(64) - F(4)$ \\
$\bullet$ Minigrid value network: $F(64) - F(5)$ \\
$\bullet$ iGibson policy network: $F(64) - F(64) - F(11)$ \\
$\bullet$ iGibson value network: $F(64) - F(64) - F(8)$

\subsection{\methodname{} in Spare Reward Settings}
\label{app:sr}
While the goal of \methodname{} is not to handle sparse rewards, we empirically evaluated the success and failure cases of our method in sparse reward settings.
We perform this experiment in a new Minigrid domain, where the state and action spaces are the same as in Appendix~\ref{sec:minidetail}, but the agent is only rewarded at the end of the episode for reaching the goal while performing a specific arm action.

We consider two settings: 1) The agent has to perform the correct arm action at the last timestep (immediate sparse reward). 2) The agent has to perform the correct arm action at the first timestep but receives the reward at the end (delayed sparse reward). With the causal detection interval $k=1$, our method infers the correct causal relations in setting 1), but misses the causal relation between the arm action and the reward in setting 2). As a result, the learned policy is optimal in setting 1) but suboptimal in setting 2). We then tested with $k=5$ and found that our method was able to identify the causal relationships and train correctly in both settings.

To summarize, even with sparse reward and $k=1$, \methodname{} can still identify the correct causal relationships between action and sparse reward if the reward is “immediate” (i.e., only depends on the last step). When the reward is not only sparse but also delayed, \methodname{} with $k=1$ can no longer find the true causal connections, but a larger $k$ ($k=5$ in this case) allows it to identify causal connections across longer time horizons and recover the true causal correlations.

\begin{table*}[t]
    \caption{HSR Gazebo experiments on a set of tasks. Each entry in the table represents the success rate, averaged over 20 trials. A trial is considered successful if the robot can get to within $d = 0.15m$ of the target without collisions. Our method is able to zero-shot transfer to the gazebo simulator and achieve better performance compared to SLQ-MPC and the baseline RL algorithms.}
    \label{tab:addrsl}
    \centering
    \resizebox{0.8\textwidth}{!}{%
    %\begin{tabular}{ccrrrrrrrrrc}
    %  Threshold $T=$
    \begin{tabular}{ccccc}
    
    \toprule
       &  \multicolumn{3}{c}{Static Goal} & Dynamic Goal \\
     \cmidrule(lr){2-4} %\cmidrule(lr){5}
      & No Obstacle & Static Obstacles & Dynamic Obstacles & Static Obstacles \\
     \midrule
     Causal MoMa (ours)  & \textbf{20/20} & \textbf{20/20} &  \textbf{19/20} & \textbf{20/20} \\
     SLQ-MPC~\cite{pankert2020perceptive} & \textbf{20/20} & 17/20 &  3/20& 9/20 \\
     Vanilla PPO~\cite{schulman2017proximal} & 17/20 & 14/20 &  14/20& 13/20 \\
     FPPO arm-base~\cite{fu2022deep} & 16/20 & 15/20 &  14/20& 15/20 \\

    \bottomrule
    \end{tabular}
    }
\end{table*}

\subsection{Evaluation in Gazebo Simulation}
\label{app:eva}

\begin{figure}[t!]
    \centering
    \includegraphics[width=0.35\textwidth]{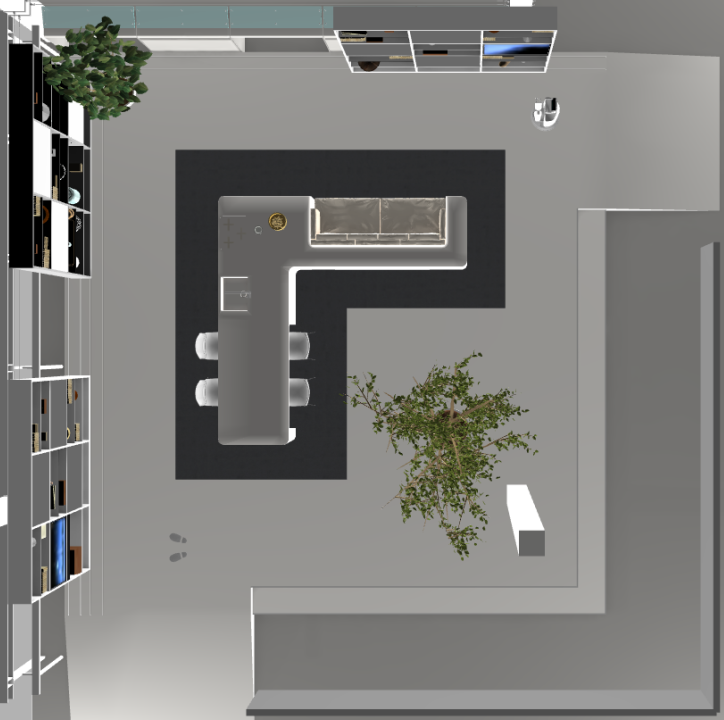}
    \caption{A top-down view of the Gazebo environment that we use to test against SLQ-MPC~\cite{pankert2020perceptive}.}
    \label{fig:gazebo}
\end{figure}

We perform empirical evaluations of \methodname{} against reactive controller by zero-shot transferring the learned policy into a Gazebo environment. 
A top-down view of our test environment is shown in Fig.~\ref{fig:gazebo}
Results are presented in Table~\ref{tab:addrsl}.

We first compare against SLQ-MPC~\cite{pankert2020perceptive}, a state-of-the-art  model predictive control (MPC) strategy for reactive whole-body mobile manipulation. Our implementation of SLQ-MPC is based on the OCS2 toolbox. We observe that a typical failure mode for SLQ is colliding with dynamic obstacles because of the lack of a mechanism to map observations to the internal model. This demonstrates that SLQ (and other similar reactive methods) rely on an explicit model of the environment to be accurate, and can often suffer when such a model is not available or inaccurate, creating additional needs for ad-hoc perceptual solutions. In addition, we found that it is non-trivial to trade-off different objectives with SLQ (e.g., obstacle avoidance, target reaching, camera angle), which results in behaviors such as converging to a local optimum behind obstacles or losing track of dynamic goals. By contrast, \methodname{} does not require building and maintaining a geometric world model and can directly map sensor signals to actions. Incorporating additional objectives into our policy is as simple as adding an extra reward term, making it easy to implement and customize. 

We also compare against the baseline learning algorithms (Vanilla PPO and FPPO arm-base) and find that our method can achieve a higher success rate in the Gazebo simulator, which corresponds to the higher reward and success rate in iGibson reflected in Fig.~\ref{fig:igibsonexp}.

\subsection{End-effector Orientation}
\label{app:eef}
\begin{figure}[ht!]
    \centering
    \includegraphics[width=0.35\textwidth]{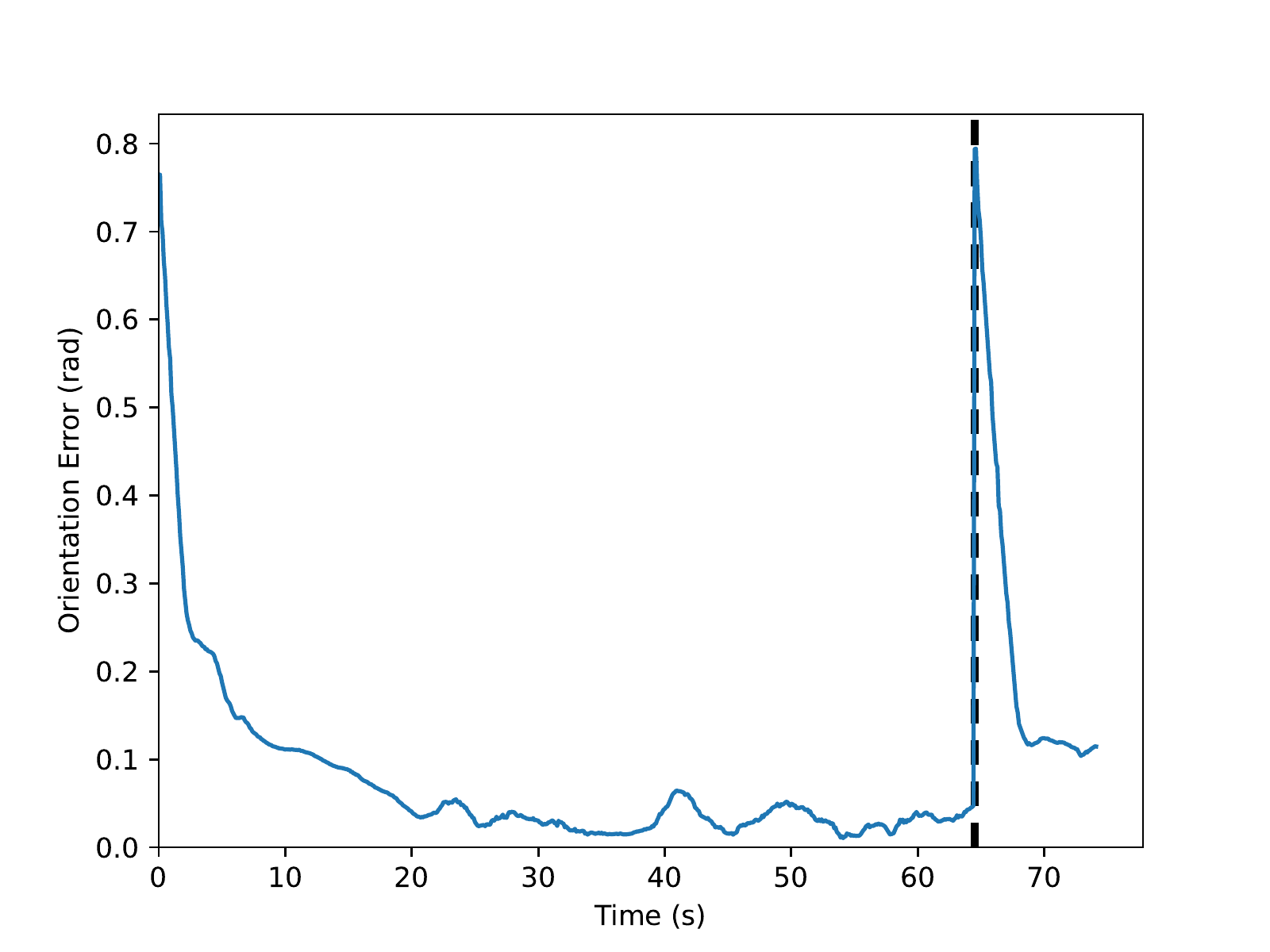}
    \caption{Orientation error over time for one of the real-world experiments with HSR. The robot is tasked with pouring a substance onto a bowl (see task in the supplementary video). Vertical dotted line: the goal orientation changes when the robot approaches the position goal but the policy trained with \methodname{} can quickly adjust to the new goal and reduce the error, which leads to a successful pouring execution of the substance (beads) into the bowl.}
    \label{fig:eef_ori}
\end{figure}

Fig.~\ref{fig:eef_ori} depicts the time evolution of the orientation error between the end-effector and the goal in one of our real-world experiments. In these experiments, we alternate dynamically between two goals for the orientation of the robot's hand, one orientation when the distance of the hand to the position goal is over \SI{1}{\meter} and a different one when it gets closer. The policy learned with \methodname{} and transferred zero-shot to the real world maintains a low error in orientation and quickly achieves the new goal when it changes (vertical line in Fig.~\ref{fig:eef_ori}), even when the scene and this dynamically changing goals are conditions never seen during training. This dynamically changing orientation resembles semantically meaningful tasks in household domains such as carrying and pouring a substance into a bowl, as illustrated in Fig.~\ref{fig:figure1} and shown in our supplementary video. Compared to the planning-based solution based on CBiRRT2, our solution keeps a mean orientation error of $\SI{0.094}{\radian}$ ($\sigma=\SI{0.13}{\radian}$) while the baseline's mean error increases to $\SI{0.73}{\radian}$ ($\sigma=\SI{0.23}{\radian}$).

\begin{figure*}[h]

\[
    B^{\text{cmi}}_{\text{Fetch}} = \kbordermatrix{
    &  R_{\text{reach}} & R_{\text{eef ori}} & R_{\text{eef height}} & R_{\text{base col}}  &  R_{\text{arm col}} &  R_{\text{self col}} &  R_{\text{head ori}} &  R_{\text{gripper}} \\
    a_{\text{forward}} & 1 & 0 & 0 & 1 & 1 & 0 & 1 & 0 \\
    a_{\text{turn}} & 1 & 0 & 0 & 1 & 1 & 0 & 0 & 0 \\
    a_{\text{pan}} & 0 & 0 & 0 & 0 & 0 & 0 & 1 & 0 \\
    a_{\text{tilt}} & 0 & 0 & 0 & 0 & 0 & 0 & 1 & 0 \\
    a_{\text{arm.x}} & 1 & 0 & 0 & 0 & 1 & 1 & 0 & 0 \\
    a_{\text{arm.y}} & 1 & 0 & 0 & 0 & 1 & 1 & 0 & 0 \\
    a_{\text{arm.z}} & 1 & 0 & 1 & 0 & 0 & 1 & 0 & 0 \\
    a_{\text{arm.rx}} & 0 & 1 & 0 & 0 & 0 & 1 & 0 & 0 \\
    a_{\text{arm.ry}} & 0 & 1 & 0 & 0 & 0 & 1 & 0 & 0 \\
    a_{\text{arm.rz}} & 0 & 1 & 0 & 0 & 0 & 1 & 0 & 0 \\
    a_{\text{grip}} & 0 & 0 & 0 & 0 & 0 & 0 & 0 & 1 
    }
\]
\[
    B^{\text{arm-base}}_{\text{Fetch}} = \kbordermatrix{
    &  R_{\text{reach}} & R_{\text{eef ori}} & R_{\text{eef height}} & R_{\text{base col}}  &  R_{\text{arm col}} &  R_{\text{self col}} &  R_{\text{head ori}} &  R_{\text{gripper}} \\
  a_{\text{forward}} & 1 & 0 & 0 & 1 & 1 & 0 & 1 & 0 \\
     a_{\text{turn}} & 1 & 0 & 0 & 1 & 1 & 0 & 1 & 0 \\
      a_{\text{pan}} & 1 & 0 & 0 & 1 & 1 & 0 & 1 & 0 \\
     a_{\text{tilt}} & 1 & 0 & 0 & 1 & 1 & 0 & 1 & 0 \\
    a_{\text{arm.x}} & 1 & 1 & 1 & 0 & 1 & 1 & 0 & 1 \\
    a_{\text{arm.y}} & 1 & 1 & 1 & 0 & 1 & 1 & 0 & 1 \\
    a_{\text{arm.z}} & 1 & 1 & 1 & 0 & 1 & 1 & 0 & 1 \\
   a_{\text{arm.rx}} & 1 & 1 & 1 & 0 & 1 & 1 & 0 & 1 \\
   a_{\text{arm.ry}} & 1 & 1 & 1 & 0 & 1 & 1 & 0 & 1 \\
   a_{\text{arm.rz}} & 1 & 1 & 1 & 0 & 1 & 1 & 0 & 1 \\
     a_{\text{grip}} & 1 & 1 & 1 & 0 & 1 & 1 & 0 & 1 
    }
\]

\caption{Causal matrices for the Fetch Robot}
\label{fig:fetchcausal}

\end{figure*}

\begin{figure*}[h]

\[
    B^{\text{cmi}}_{\text{HSR}} = \kbordermatrix{
    &  R_{\text{reach}} & R_{\text{eef ori}} & R_{\text{eef height}} & R_{\text{base col}}  &  R_{\text{arm col}} &  R_{\text{self col}} &  R_{\text{head ori}} &  R_{\text{gripper}} \\
    a_{\text{forward}}    & 1 & 0 & 0 & 1 & 1 & 0 & 0 & 0 \\
    a_{\text{side}}       & 1 & 0 & 0 & 1 & 1 & 0 & 0 & 0 \\
    a_{\text{turn}}       & 1 & 0 & 0 & 0 & 1 & 0 & 1 & 0 \\
    a_{\text{pan}}        & 0 & 0 & 0 & 0 & 0 & 0 & 1 & 0 \\
    a_{\text{tilt}}       & 0 & 0 & 0 & 0 & 0 & 0 & 1 & 0 \\
    a_{\text{arm.lift}}   & 1 & 0 & 1 & 0 & 1 & 1 & 1 & 0 \\
    a_{\text{arm.flex}}   & 1 & 1 & 1 & 0 & 1 & 1 & 0 & 0 \\
    a_{\text{arm.roll}}   & 0 & 1 & 0 & 0 & 0 & 1 & 0 & 0 \\
    a_{\text{wrist.flex}} & 0 & 1 & 0 & 0 & 0 & 1 & 0 & 0 \\
    a_{\text{wrist.roll}} & 0 & 1 & 0 & 0 & 0 & 1 & 0 & 0 \\
    a_{\text{grip}}       & 0 & 0 & 0 & 0 & 0 & 0 & 0 & 1 
    }
\]

\[
    B^{\text{arm-base}}_{\text{HSR}} = \kbordermatrix{
    &  R_{\text{reach}} & R_{\text{eef ori}} & R_{\text{eef height}} & R_{\text{base col}}  &  R_{\text{arm col}} &  R_{\text{self col}} &  R_{\text{head ori}} &  R_{\text{gripper}} \\
    a_{\text{forward}}    & 1 & 0 & 0 & 1 & 1 & 0 & 1 & 0 \\
    a_{\text{side}}       & 1 & 0 & 0 & 1 & 1 & 0 & 1 & 0 \\
    a_{\text{turn}}       & 1 & 0 & 0 & 1 & 1 & 0 & 1 & 0 \\
    a_{\text{pan}}        & 1 & 0 & 0 & 1 & 1 & 0 & 1 & 0 \\
    a_{\text{tilt}}       & 1 & 0 & 0 & 1 & 1 & 0 & 1 & 0 \\
    a_{\text{arm.lift}}   & 1 & 1 & 1 & 0 & 1 & 1 & 1 & 1 \\
    a_{\text{arm.flex}}   & 1 & 1 & 1 & 0 & 1 & 1 & 1 & 1 \\
    a_{\text{arm.roll}}   & 1 & 1 & 1 & 0 & 1 & 1 & 1 & 1 \\
    a_{\text{wrist.flex}} & 1 & 1 & 1 & 0 & 1 & 1 & 1 & 1 \\
    a_{\text{wrist.roll}} & 1 & 1 & 1 & 0 & 1 & 1 & 1 & 1 \\
    a_{\text{grip}}       & 1 & 1 & 1 & 0 & 1 & 1 & 1 & 1 
    }
\]
\caption{Causal matrices for the HSR Robot}
\label{fig:hsrcausal}
\end{figure*}

\end{document}